\documentclass[10pt,twocolumn,twoside]{IEEEtran}
\usepackage{graphicx}
\usepackage{amsmath}
\usepackage{amssymb}
\usepackage[tight]{subfigure}
\usepackage{cite}
\usepackage{bm}
\usepackage{multirow}
\usepackage{booktabs}
\usepackage{makecell}
\usepackage{algpseudocode}
\usepackage{color}
\usepackage[dvipsnames]{xcolor}
\usepackage[export]{adjustbox}
\usepackage[pagebackref=true,breaklinks=true,colorlinks,bookmarks=false,linkcolor=blue,citecolor=blue]{hyperref}
\begin{document}

\title{Adaptive Blind Super-Resolution Network for
Spatial-Specific and Spatial-Agnostic Degradations}

\author{Weilei~Wen, Chunle~Guo, Wenqi~Ren, Hongpeng~Wang, and Xiuli~Shao
	\thanks{W. Wen, C. Guo, and X. Shao are with the VCIP, College of Computer Science, Nankai University, Tianjin, 300350, China (e-mail: wlwen@mail.nankai.edu.cn). 
 
    W. Ren is with the School of Cyber Science and Technology, Shenzhen Campus, Sun Yat-sen University, Shenzhen 518107, P.R. China (e-mail: rwq.renwenqi@gmail.com). 
 
    H. Wang is with the College of Artificial Intelligence, Nankai University, Tianjin, 300350, China, and China and Laboratory of Science and Technology on Integrated Logistics Support, National University of Defense Technology, Changsha 410073, China (e-mail: hpwang@nankai.edu.cn).

    Corresponding author: H. Wang (e-mail: hpwang@nankai.edu.cn)

}
}

\markboth{IEEE Transactions on Image Processing}{}

\maketitle

\begin{abstract}
Prior methodologies have disregarded the diversities among distinct degradation types during image reconstruction, employing a uniform network model to handle multiple deteriorations. Nevertheless, we discover that prevalent degradation modalities, including sampling, blurring, and noise, can be roughly categorized into two classes.
We classify the first class as spatial-agnostic dominant degradations, less affected by regional changes in image space, such as downsampling and noise degradation.
The second class degradation type is intimately associated with the spatial position of the image, such as blurring, and we identify them as spatial-specific dominant degradations.
We introduce a dynamic filter network integrating global and local branches to address these two degradation types. This network can greatly alleviate the practical degradation problem. 
Specifically, the global dynamic filtering layer can perceive the spatial-agnostic dominant degradation in different images by applying weights generated by the attention mechanism to multiple parallel standard convolution kernels, enhancing the network's representation ability.
Meanwhile, the local dynamic filtering layer converts feature maps of the image into a spatially specific dynamic filtering operator, which performs spatially specific convolution operations on the image features to handle spatial-specific dominant degradations. By effectively integrating both global and local dynamic filtering operators, our proposed method outperforms state-of-the-art blind super-resolution algorithms in both synthetic and real image datasets.

\end{abstract}


\IEEEpeerreviewmaketitle
%% main text

\section{Introduction}
\label{sec-intro}
Single image super-resolution (SISR) is a crucial computer vision task that focuses on reconstructing high-resolution (HR) images from low-resolution (LR) counterparts. It finds extensive applications in image restoration and medical imaging domains. 
In recent years, convolutional neural network-based SR methods~\cite{dong2014learning, kim2016deeply, lim2017enhanced, zhang2018learning, niu2020single} have made significant progress with the development of deep learning technology. Researchers have achieved promising reconstruction results on synthetic datasets by enhancing network representational capability.

However, practical degradation typically involves multiple factors, such as blur, noise, and down-sampling, far beyond the simple down-sampling assumption. Although methods like~\cite{zhang2018learning, niu2020single, dai2019second, he2016deep} have shown promising results on synthetic benchmark datasets, their performance severely deteriorates when faced with real applications. Therefore, significant challenges in solving the practical degradations image super-resolution problem still exist.
\begin{figure}[t]\footnotesize
	\begin{center}
		\tabcolsep 1.5pt
		\includegraphics[width = 0.45\textwidth]{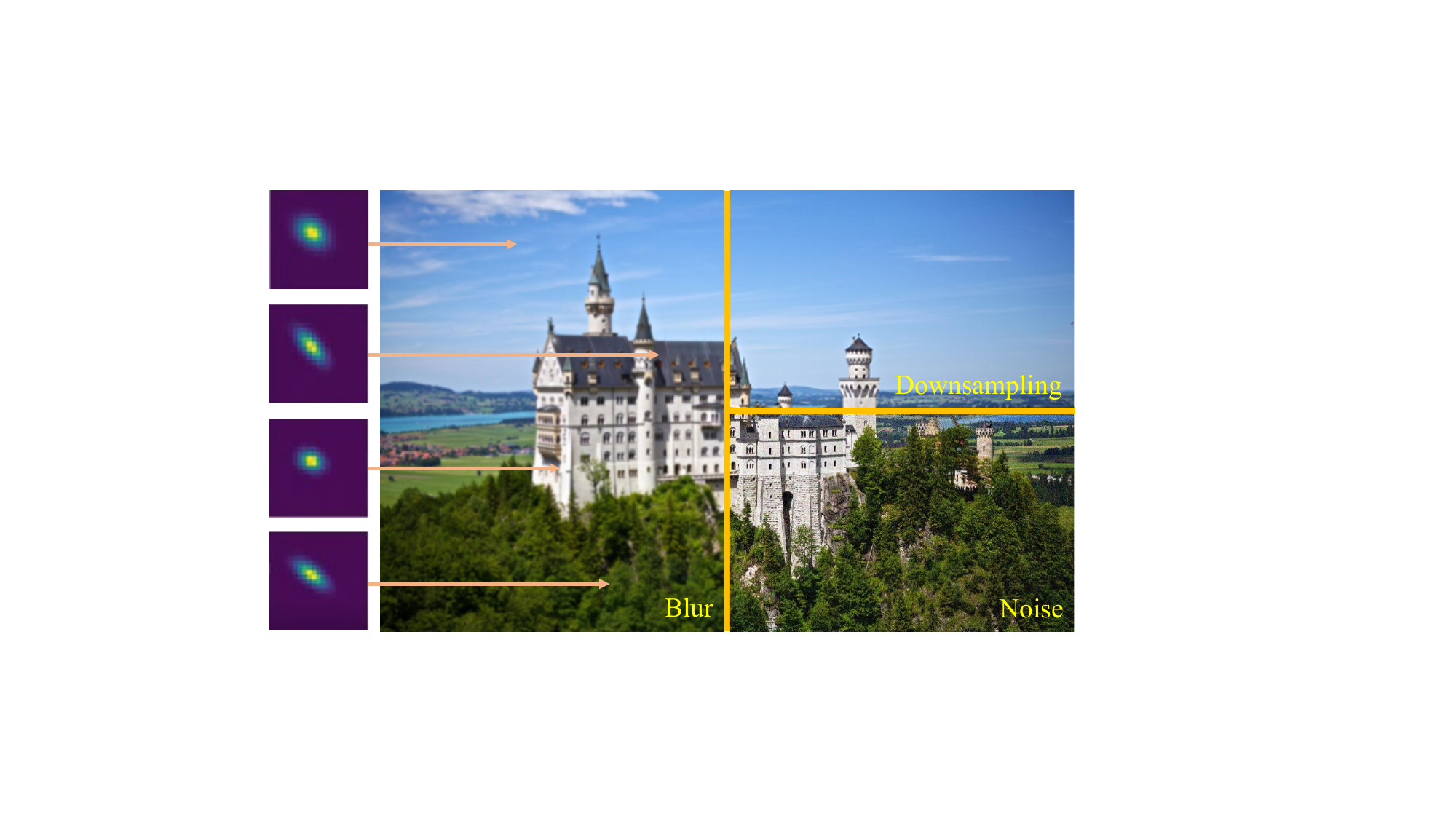}
	\end{center}
	\caption{This image is $``img\_87.png"$ from the DIV2KRK~\cite{bell2019blind} dataset. The image is divided into three parts containing different degradation effects: blur, downsampling, and noise. On the left side of the image are four blur kernels, which are estimated from different positions in the blurred area.  } 
	\label{fig: degradation}
\end{figure}

To address complex degradation problems in practical applications, researchers have proposed blind super-resolution (BSR) algorithms. Compared to non-blind SISR methods, BSR algorithms can better perceive multiple degradations in images and show more promising potential for real applications.
Unlike non-blind super-resolution, which only uses down-sampling of HR images to create LR counterparts, blind super-resolution requires additional steps like blurring and adding noise to get the training data pairs.
Specifically, in blind SR methods, the LR image $y$ is obtained from the HR image $x$ through the following degradation model:
\begin{gather}
	y = (x\circledast k)\downarrow_{s} + n,
	\label{eqn: blind}
\end{gather}
First, the HR image obtains the blurred image by the convolution operation $\circledast$ with the blur kernel $k$. Subsequently, the blurred image reduces the image resolution by downsampling operation $\downarrow_{s}$ with factor $s$ and introducing additive noise $n$ to obtain the final LR image $y$. According to previous work~\cite{luo2022deep, luo2021endtoend, luo2020unfolding}, the blur kernel usually contains various isotropic and anisotropic Gaussian blur kernels, the downsampling method adopts the s-fold downsampling operation and $n$ is Additive White Gaussian Noise with a noise level of $\sigma$.

To solve multiple degradation problems in images, some blind super-resolution methods~\cite{gu2019blind,luo2021endtoend,luo2020unfolding,luo2022deep,shi2020ddet} introduce a blur kernel estimation branch to simulate the blur information contained in degraded images and use it as prior knowledge to achieve the goal of deblurring. In \cite{bell2019blind}, Michal \emph{et al.} utilized the degradation distribution similarity of image cross-scale patches to estimate the blur kernel by KernelGAN. Providing the estimation kernel for non-blind SR methods can significantly improve the reconstruction performance. 
Gu \emph{et al.} \cite{gu2019blind} proposed a blind SR method that contains a kernel estimation network and an SR network. They introduced an iterative kernel correction (IKC) optimization algorithm to correct estimation kernel errors. Although this method can improve the accuracy of kernel estimation, it is time-consuming and computationally expensive.
Luo \emph{et al.} \cite{luo2020unfolding} designed two modules, namely restorer and estimator. The restorer reconstructs the SR image based on the predicted blur kernel, while the estimator estimates the blur kernel through the restored SR image. These two modules run alternately to restore and reconstruct degraded images.

In dealing with multiple degradations, the above BSR methods usually use a unified network to address all degradation types. However, some degradation types may appear spatial-specific dominant among common degradations, while others manifest as spatial-agnostic dominant. Hence, employing a single-branch or convolution-based dual-branch network to address multiple degradation issues may not be optimal.
As shown in Figure~\ref{fig: degradation}, the image undergoes three degradation processes: blur, noise, and bicubic. Blur kernels typically tend to be spatial-specific. 
Practical degradation images present different blurring, including motion blur and defocus blur, due to the comprehensive influence of various factors such as object motion, depth of field difference, out-of-focus conditions during shooting, and camera shake. These factors cause the image to exhibit different types and degrees of blur at different locations, for example, the difference between foreground and background. 
Unlike blur, noise and downsampling in an image are usually spatial-agnostic. Common additive white Gaussian noise demonstrates no correlation with the signal strength of the image. Typically applied uniformly to the entire image during the noise addition process, it lacks spatial specificity. Downsampling processes images through specific algorithms that present spatial-agnostic dominant characteristics. 

In addition, the previous method usually has the following problems: i) Estimated blur kernels may be inaccurate and incompatible with SR networks, affecting restoration results. Moreover, BSR algorithms with kernel estimation lack generalization ability for different types of degradation (e.g., noise). ii) In previous models, the SR model and kernel estimation module used standard convolution as the basic unit. However, images usually contain various spatial-specific degradations in practical applications, so forcing the entire image to be processed with the same convolution kernel is unreasonable. Besides, some methods~\cite{li2021efficient,liang2021swinir,chen2023activating} employ the Transformer architecture to solve the BSR or classical SR problem. Although the Transformer's self-attention mechanism can capture long-range dependencies in images, focus on different regions of the image, and understand the global contextual information of the image to better address spatially varying degradation problems, such methods usually have high computational complexity and memory requirements when dealing with high-resolution images.

Drawing upon the analysis above, we propose a dual-branch dynamic filter network to address the previously mentioned issues. Specifically, inspired by dynamic filter networks and their various variants~\cite{chen2020dynamic,zhou2021decoupled}, we proposed a novel architecture, the Global and Local Dynamic Filter Network (GLDFN), which uses dynamic filter convolution as the basic unit. The proposed GLDFN includes the local and global dynamic filter branches, which aim to tackle spatial-specific dominant and spatial-agnostic dominant degradations. The local dynamic filtering layers employ distinct filter kernels for image pixel blocks in various regions, which aids in resolving the challenges posed by spatial-specific dominant degradations in practical scenes. The global dynamic filtering layer augments the network's representative capacity and can distinguish spatial-agnostic dominant degradation between different samples.
The dynamic filtering network possesses two inherent advantages in addressing blind super-resolution problems. Firstly, from a content perspective, different regions of an image contain distinct information, and utilizing dynamic filtering networks enables a more discriminative treatment of different content information, thereby facilitating the more accurate and flexible extraction of feature information. Secondly, from the perspective of degradation patterns, different regions of an image contain different degradation patterns. Dynamic filtering network generates different weights, which can perceive various degradation types, thus possessing an advantage in handling realistic images.

This paper presents a novel blind super-resolution network called GLDFN to effectively solve the problem of spatial-specific dominant and spatial-agnostic dominant degradations in images. Our contributions can be summarized as follows:
\begin{itemize}
	\item[\textbullet] We divide the practical degradations into two categories: spatial-specific dominant degradations and spatial-agnostic dominant degradations. Then, we propose a dynamic filter-based blind SR method, namely the Global and Local Dynamic Filter Network (GLDFN), to address these two types of degradation. Our method can better perceive multiple degradations and enhance feature representation, thereby improving the ability of super-resolution reconstruction.
	\vspace{0.5mm}
	\item[\textbullet] We introduce global and local dynamic filtering to learn the representation of spatial-agnostic dominant and spatial-specific dominant degradations, respectively. Through the collaboration between the global and local dynamic filtering layers, our method effectively restores images with various practical degradations. 
\end{itemize}

We perform quantitative and qualitative evaluations on various benchmark datasets and the experimental results show that our method outperforms the state-of-the-art blind SR methods.
\begin{figure*}[t]
	\begin{center}
		\includegraphics[width=0.98\linewidth]{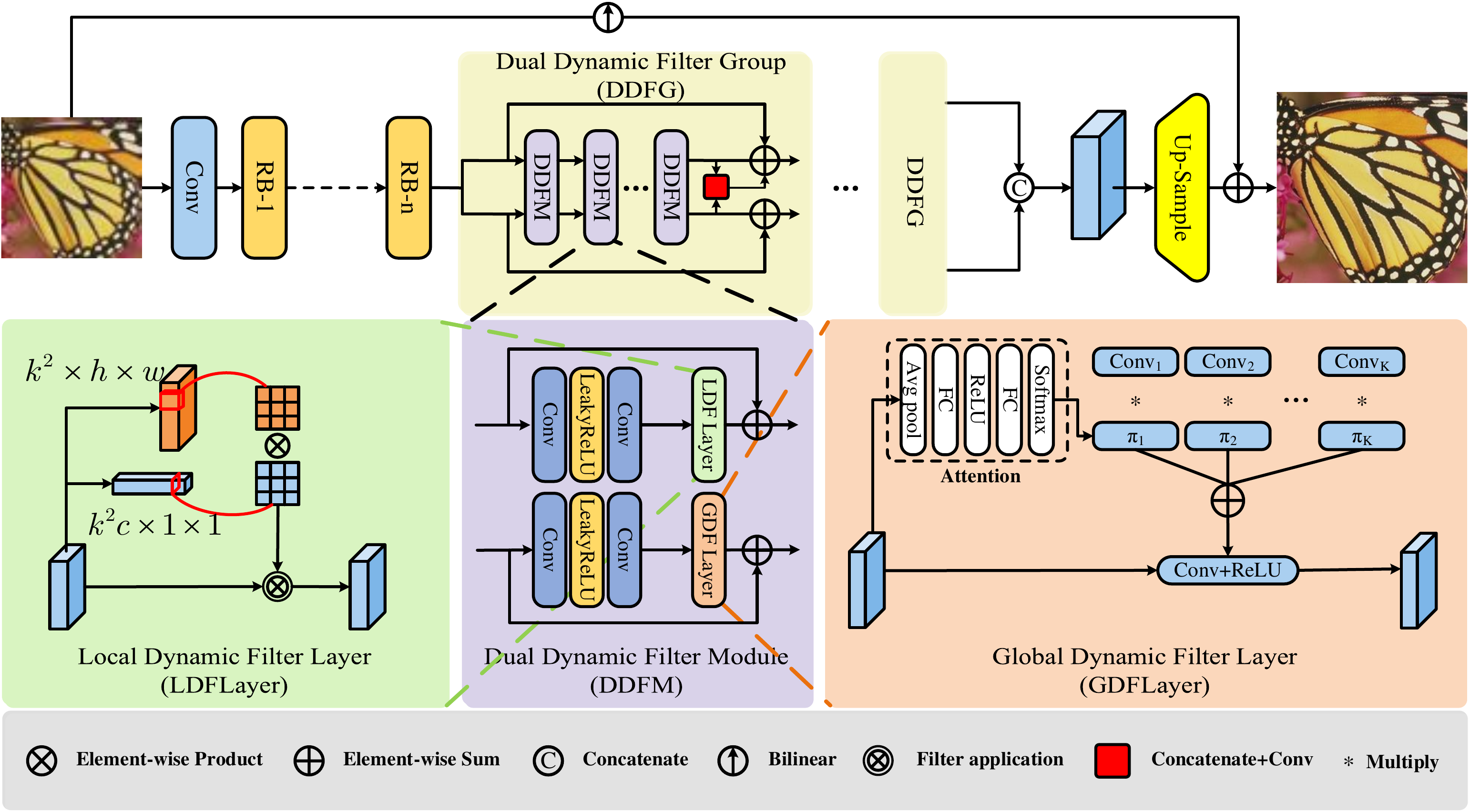}
	\end{center}
	\caption{The architecture of the proposed global and local dynamic filters network (GLDFN). Given an LR image input, we obtain shallow blur features from it. Then, the two types of degradations will be cleaned through the dual dynamic filters refinement network, which consists of Dual Dynamic Filter Groups (DDFG). Each DDFG encompasses multiple Dual Dynamic Filter Modules (DDFM). These modules consist of local and global branches, comprising sequences of convolutional layers and local and global dynamic filter layers. Finally, the Reconstruction Network recovers a high-resolution image from cleaned features.}
	\label{fig: architecture}
\end{figure*}

\section{Related Work}
\label{sec-related}
This section reviews the related work on single image super-resolution (SISR) and dynamic filter networks. In more detail, SISR is divided into two categories: single degradation image SR and complex degradations image SR.

\subsection{Single Image Super-Resolution}
% \vspace{4pt}
\noindent \textbf{SR with Single Degradation. }Dong et al.~\cite{dong2014learning} proposed a groundbreaking CNN-based SR architecture named SRCNN, which laid the foundation for deep learning applied to SISR methods. Since then, deep learning-based SR algorithms have had a booming development. Kim et al.~\cite{kim2016accurate} deepened the SR network to 20 layers through residual learning techniques, and Lim et al.~\cite{lim2017enhanced} established a deep SR model named EDSR using modified ResNet~\cite{he2016deep} residual blocks. To further improve the perceptual quality, adversarial learning is adopted to restore more realistic results in~\cite{wang2021real,wang2018esrgan,zhang2019ranksrgan}. 
Recently, attention mechanisms have been introduced into SR methods to enhance the network's feature representation ability. For example, RCAN~\cite{zhang2018image} and SAN~\cite{dai2019second} proposed channel attention and second-order channel attention to exploit feature correlations. HAN~\cite{niu2020single} introduced layer attention and channel-spatial attention modules to extract informative features and improve network performance. 
Recent research has optimized SISR techniques by employing binarized neural networks~\cite{xin2023advanced,xin2020binarized}, and wavelet-based networks~\cite{xin2020wavelet} to improve the models' computational efficiency and inference speed while maintaining or improving the image reconstruction quality.
More recently, vision transformer has been widely applied to various computer vision tasks, and many transformer-based methods~\cite{chen2021pre,li2021efficient,liang2021swinir,chen2023activating,zhang2024eatformer} continue to break the highest records in the SR task. Vision transformer demonstrates tremendous advantages in feature representation.

\vspace{4pt}
\noindent \textbf{SR with Multiple Degradations. }The above SR methods tend to settle super-resolution problems with a single degradation setting (\emph{e.g.} bicubic downsampling), which differs from practical degradations. These methods handling a single type of degradation will suffer a serious performance drop in real applications. Therefore, several blind SR methods \cite{zhang2018learning,xu2020unified,zhang2020deep,hussein2020correction} have been proposed to handle complex degradation SR problems. 
%non-blind sr with multiple degradations
%
Specifically, SRMD \cite{zhang2018learning} adopted degradation maps and LR images as input to handle multiple degradation SR applications. After that, UDVD~\cite{xu2020unified} utilized standard convolution to extract features and gradually employed dynamic convolution to enhance the sharpness of the image. However, this method indiscriminately applies dynamic convolution to multi-degradation super-resolution tasks without distinguishing between spatial-specific and spatial-agnostic degradations. 
Zhang \emph{et al.} \cite{zhang2020deep} designed an end-to-end unfolding network that leverages learning- and model-based methods to benefit practical image restoration. 
The KDSR~\cite{xia2023knowledge} and MRDA~\cite{xia2023meta}  methods utilize knowledge distillation techniques to enable the student network to extract the same implicit representation of degradation as the teacher's network directly from the LR image and use this degradation representation to guide the generation of SR. However, this degradation estimation method may be unstable because it is difficult to guarantee that the student network can fully learn the same degradation feature extraction capability as the teacher network.

Furthermore, zero-shot techniques~\cite{shocher2018zero,soh2020meta} have also been studied in non-blind SR with multiple types of degradation. Among them, ZSSR~\cite{shocher2018zero} is one of the most representative models, which first attempted to train a self-supervision network based on a patch recurrence strategy. This method can super-resolve LR images without a pre-training step. 

Since precise blur kernels are helpful for blind SR, the accuracy of kernel estimation will be the bottleneck of SR performance. To address this problem, KernelGAN \cite{bell2019blind} introduced an internal generative adversarial network(GAN) to estimate the blur kernel and utilize that kernel to assist a non-blind SR approach for performance improvement.
Gu \emph{et al.} \cite{gu2019blind} proposed an iterative kernel correction (IKC) method, which can correct the estimation kernel error iteratively.
Similarly, IKR-Net~\cite{ates2023deep} introduces an iterative kernel BSR network that includes blur kernel and noise estimation. This network achieves high-quality reconstruction for noisy images by iteratively refining the reconstructed image and the estimated blur kernel.
Luo \emph{et al.} \cite{luo2020unfolding,luo2021endtoend,luo2023end} implemented SR recovery by designing a restorer module and an estimator module. The recoverer module recovers the SR image based on the estimated degradation, while the estimator module utilizes the recovered SR image to estimate the degradation. By alternately optimizing these two modules, these methods form an end-to-end trainable blind super-resolution network, which addresses the compatibility problem between the SR model and kernel estimation model. However, these approaches can only estimate a single kernel and cannot handle spatially varying degradations.
Inspired by contrastive learning, Wang \emph{et al.} \cite{wang2021unsupervised} introduced a degradation-aware SR model, which can flexibly adapt to various degradations.
Likewise, DSAT~\cite{liu2024degradation}integrates contrastive learning to capture the image degradation representation. DASR~\cite{liang2022efficient} predicts the degradation parameters of the input image using a small regression network and co-optimizes the network parameters with multiple convolutional experts sharing the same topology to handle varying degradation levels. Yang \emph{et al.}~\cite{yang2024dynamic} proposed a Dynamic Kernel Prior (DKP) model to realize real-time kernel estimation and solve the BSR problem using an unsupervised paradigm.
KULNet~\cite{fang2022uncertainty} adopted an uncertainty learning scheme to enhance the robustness of the kernel estimation network.

Several recent studies~\cite{laroche2023deep,kim2021koalanet,liang2021mutual,cornillere2019blind,xu2021edpn,zhou2022joint} have considered spatially variant blur into blind super-resolution.  
DMBSR~\cite{laroche2023deep} proposes a fast, deep plug-and-play algorithm based on a linearized ADMM splitting technique. This innovative approach solves the SR problem caused by spatial variation blur.
KOALAnet~\cite{kim2021koalanet} introduces a BSR network that leverages kernel-oriented adaptive local adjustment to learn the spatially variant degradation types. However, this method requires greater complexity and time consumption during training than the end-to-end approaches.
MANet~\cite{liang2021mutual} designed a moderate receptive field convolution layer to better adapt to the local degradation characteristics.
Blind SR~\cite{cornillere2019blind} integrated a degradation-aware SR network and a kernel discriminator to estimate the blur kernel, thereby facilitating more accurate blur estimation.

\subsection{Dynamic Filter Networks}
The convolutional layer is a foundational element in computer vision tasks. Within image processing, convolution showcases the trait of weight sharing, wherein distinct positions share identical weights. Although this sharing improves efficiency, it can cause the convolutional layer to lack sensitivity to image content and hinder its efficacy in addressing the diverse degradation images. In Sec.~\ref{sec-intro}, we categorize practical degradations into two classes through analysis: spatial-agnostic dominant and spatial-specific dominant degradations. The convolutional layer, which relies on spatially shared weights, may not be optimally suited for addressing multi-degradation problems in imagesThe convolutional layer cannot distinguish various degradations within an image, i.e., spatial-specific and spatial-agnostic degradations. To overcome this problem, we introduce dynamic filter layers to address these practical degradations. Subsequently, we provide a brief overview of the evolution of the dynamic convolutional network.

Jia \emph{et al.} \cite{jia2016dynamic} first attempted to investigate the dynamic filter network.
After that, a dynamic filtering network typically consists of two or more branches, one of which is the identity map of the original input, and the other branches are responsible for generating dynamic filtering kernels at each pixel or predicting convolutional kernel weights.
Dynamic filters~\cite{adaptive_filter,zhou2021decoupled,solov2,cond_seg,magid2021dynamic} leverage distinct network branches to predict an individual filter at each pixel. Because each generated dynamic filter kernel is region-specific for the feature, it can better perceive the degradation information of different feature regions.
However, it is important to note that the resulting combined filters in~\cite{condconv,weightnet,dynet,chen2020dynamic} are still applied in a convolutional manner, where they maintain spatial sharing within the same input.

In this work, we adopt the dynamic filter into our method and build a global and local dynamic filter network for addressing the multi-degradation problem. For spatial-specific dominant degradations, we introduce so-called local dynamic filtering layers. These filter layers generate distinct filters within various image or feature regions, enhancing the ability to perceive spatially specific degradation types. Different degradations in distinct areas can be flexibly handled using adaptive local dynamic filters. We also introduce a dynamic filtering layer termed ``global" for spatial-agnostic dominant degradations. Unlike the local dynamic filtering layer, it does not generate local convolution kernels for images or features. Instead, the global dynamic filtering layer generates varied convolution kernel weights for different samples. On the one hand, these global dynamic filtering layers enhance the network's representative capacity. On the other hand, since the convolution kernel weights are input-dependent, they can discriminate the spatial-agnostic dominant degradations among different samples. With extensive experiments, we have verified that dynamic filters can be successfully used in blind SR tasks to solve various degradation problems flexibly.
\begin{figure*}[t]
	\setlength{\abovecaptionskip}{-0.05in}
	\setlength{\belowcaptionskip}{-0.05in}
	\begin{center}
		\includegraphics[width=.999\linewidth]{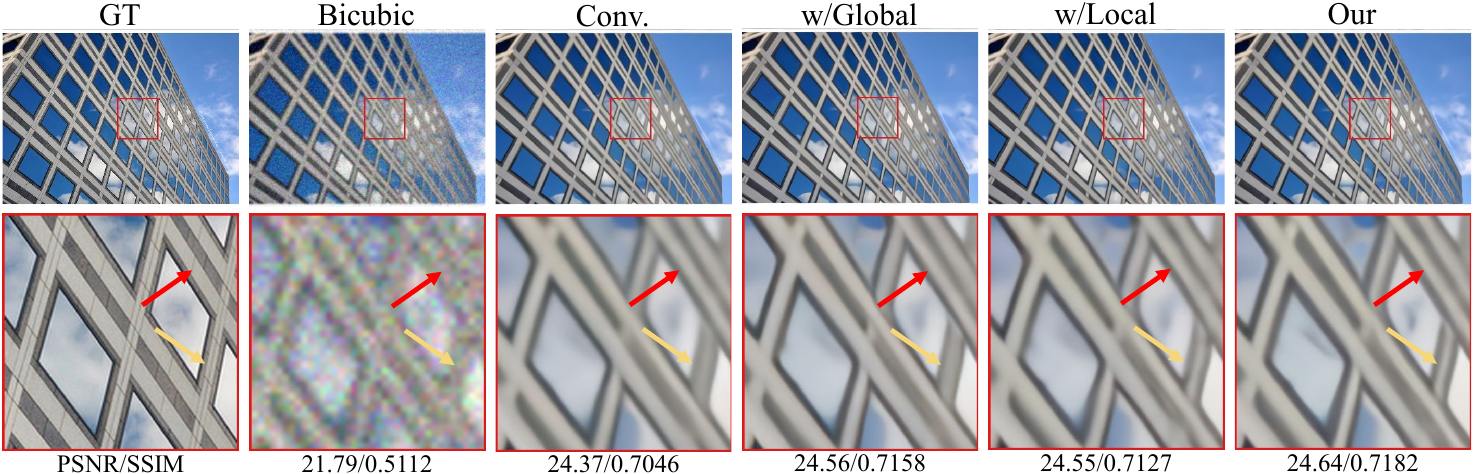}
	\end{center}
	\caption{{Ablation experiments for 4$\times$ SR with different network architectures on the Urban100~\cite{huang2015single} dataset. Our algorithm GLDFN with all the proposed modules outperforms other configurations.}}
	\label{fig: ablation}
\end{figure*}

\begin{table*}[t]
    \small
    \centering
    \setlength\tabcolsep{7pt}
    \caption{Ablation studies of network architectures. Results are reported as average PSNR and SSIM on three datasets:
Urban100~\cite{huang2015single} for Setting 1, DIV2KRK~\cite{bell2019blind} for Setting 2, and $COCO\_Valid\_200$~\cite{lin2014microsoft} for Setting 3.}
    \begin{tabular}{cccccccccccc}
        \toprule[0.15em]
            
\multirow{3}{*}{Exp.} & \multirow{3}{*}{Global} & \multirow{3}{*}{Local} & \multicolumn{2}{c}{Setting1} & \multicolumn{2}{c}{Setting2} &\multicolumn{2}{c}{Setting3} & \multirow{3}{*}{Params (M)} \\
                     &              &                &\multicolumn{2}{c}{Urban100}          &\multicolumn{2}{c}{DIV2KRK}     &\multicolumn{2}{c}{COCO }   & \\     
                     &              &                &PSNR&SSIM         &PSNR&SSIM     &PSNR&SSIM   &  \\ 
        \midrule[0.15em]
        A        &			    &                & 23.05&0.6423             & 32.53&0.9017     & 22.75&0.6231          & 14.736 \\ 
            B        & $\checkmark$ &                & 23.13&0.6482          & 32.65&0.9045   & 22.80&0.6249            & 14.667 \\
        C        &              & $\checkmark$   & 23.17&0.6484                  & 32.62&0.9037        &22.82&0.6267         & 14.695 \\
        D		 & $\checkmark$ & $\checkmark$   & \textbf{23.20}&\textbf{0.6521}  &\textbf{32.76}&\textbf{0.9062}  & \textbf{22.86}&\textbf{0.6280}  & 14.696 \\ 
        \bottomrule[0.1em]
    \end{tabular}
    \label{tab: ablation}
\end{table*}

\section{Our Method}
\label{sec-method}
Figure \ref{fig: architecture} illustrates the proposed global and local dynamic filters network (GLDFN) architecture. In this section, we outline the proposed method's pipeline.

\subsection{Network Overview}
As depicted in Figure \ref{fig: architecture}, we introduce a network GLDFN based on dynamic filters to tackle the spatial-agnostic dominant and spatial-specific dominant degradations problem. 
The GLDFN comprises three primary components: the Shallow Feature Extraction Network, the Dual Dynamic Filters Refinement Network, and the Reconstruction Network.

In contrast to previous methods that explicitly formulate a blur kernel estimation branch, we achieve greater flexibility and efficiency in addressing spatial-specific and spatial-agnostic degradations by utilizing global and local dynamic filters. Given a low-resolution (LR) image $I^{LR}$, we extract shallow blurred features through the shallow feature extraction network composed of modified residual blocks. Subsequently, these features are fed to the Dual Dynamic Filter Modules' (DDFMs) local and global branches to handle distinct degradation types independently. Using the Residual in Residual (RIR) structure, we combine multiple DDFMs to form a Dual Dynamic Filter Group (DDFG) to improve the representation ability of the network.
We refine blurred features through multiple successive DDFGs and obtain clear image features. The two branches' features are merged in the reconstruction module and directed to the up-sampling module for image reconstruction. Finally, we recover an HR image from its LR counterpart.

\subsection{Shallow Feature Extraction Network}

We consider an LR image $I^{LR}$ with multiple degradations as input. The input undergoes an initial process through a convolution layer, generating a shallow feature $F_{0}$.
\begin{equation}
    F_{0} = \operatorname{Conv}(I^{LR}),
\end{equation}

To further improve the representation ability of our network, we exploit $n$ modified residual blocks ($\operatorname{RB}$) to extract features $F_{i}$. Specifically, each $\operatorname{RB}$ comprises three convolutions, with all the layers, except for the last convolution layer, succeeded by nonlinear activation.
\begin{equation}
	F_i = \operatorname{RB}_i(F_{i-1}), \quad i = 1, 2, \ldots, n,
\end{equation}
where $F_{i-1}$ is the input of the $i$-th residual block $\operatorname{RB}_{i}$.

\subsection{Dual Dynamic Filters Refinement Network}

Previous blind SR algorithms commonly rely on standard convolutional operations to construct the blur kernel estimation module. However, the weight-sharing nature of convolutions may not be conducive to solving spatial-specific dominant and spatial-agnostic dominant degradations.
Accordingly, we introduce a dual dynamic filtering refinement network, consisting of multiple DDFGs, to effectively tackle spatial-specific and spatial-agnostic degradations. Each DDFG incorporates multiple DDFMs. As illustrated in Figure~\ref{fig: architecture}, the DDFM consists of global and local dynamic filtering branches.
The local dynamic filter layer produces distinct filters in various regions of the image or feature, enabling adaptable handling of spatial-specific degradations in different areas. Meanwhile, the global dynamic filter layer generates distinct convolution kernel weights for separate samples, aiding in perceiving spatial-agnostic dominant degradations across various samples.

\vspace{4pt}
\noindent \textbf{Dual Dynamic Filter Group. }We employ the RIR structure, as proposed in~\cite{zhang2018image}, to construct the dual dynamic filter groups (DDFGs). Each DDFG consists of multiple dual dynamic filter modules (DDFMs).
Additionally, a long skip connection is imposed on each group, promoting training stability while enhancing the network's representation and reconstruction ability. Following each dual dynamic filter group, we employ a fusion module to merge the outcomes of the two branches, thus facilitating a more harmonized handling of the diverse degradation within the image. In each DDFM, we constructed a dual path structure using local and global dynamic filter branches, as shown in the DDFM in Figure \ref{fig: architecture}.
Within the DDFM, instead of directly connecting the two branches, we let them work independently to improve their respective ability to address distinct forms of degradation.
Before the local and global dynamic filter layers, we introduce two convolutional layers and an activation layer. Like the residual block, we incorporate a skip connection to stabilize the training process.

We are the first to propose a dual-branch structure incorporating two distinct dynamic filters. This dual dynamic filter architecture offers a superior alternative to traditional convolutional layers when dealing with both spatial-agnostic dominant and spatial-specific dominant degradations. The local dynamic filter layer is designed to generate position-dependent filters across the image or feature space, thereby allowing for adaptive handling of region-specific degradation. Concurrently, the global dynamic filter layer produces sample-specific convolution kernel weights, which enhances the network's capability to accurately discern and address the spatial-agnostic dominant degradation across various samples.
 
\vspace{4pt}
\noindent \textbf{Global Dynamic Filter Layer. }As discussed in Sec.~\ref{sec-intro}, real-world low-resolution images often exhibit a mix of degradation types, broadly classified into spatial-agnostic degradations (e.g., downsampling) and spatial-specific degradations (e.g., blur). Our network architecture employs a global dynamic filter layer to manage the former, while the local dynamic filter layer is dedicated to handling the latter. This dual dynamic filter network ensures a comprehensive and effective resolution of the complex degradations present in practical scenarios.

We incorporate the concept of dynamic convolution~\cite{chen2020dynamic} into our architecture to build the global dynamic filter layer. As shown in the Global Dynamic Filter Layer (GDFLayer) in Figure \ref{fig: architecture}, this layer employs an attention mechanism to generate a sequence of weights to modulate each convolutional layer's kernels. 
On the one hand, since the generated convolution kernel weights are related to the input, for different images, the global dynamic filter layer will generate different convolution kernels for processing. This sample-specific convolution kernel is beneficial for the network to perceive the spatial-agnostic dominant degradations in different input samples. On the other hand, generating input-dependent weights for the convolution kernel enhances its feature selection characteristics, resulting in more comprehensive feature information extraction. Next, we briefly review the mechanism of the GDFLayer as follows.

We define the traditional perceptron as $\bm{y}=f(\bm{W}^T\bm{x}+\bm{b})$, where $\bm{x}$, $\bm{y}$, $\bm{W}$, and $\bm{b}$ represent the input (e.g., image), output, weight matrix, and bias vector, and $f$ is an activation function (such as ReLU~\cite{NairH10Relu, JarrettKRL09Relu}). The dynamic perceptron is formed by aggregating multiple (K) linear functions $\bm{\tilde{W}}^T_k\bm{x}+\bm{\tilde{b}}_k$, defined as follows:
\begin{align}
	% \bm{y} = &f\left(\sum_{k=1}^K\pi_k(\bm{x})(\bm{\tilde{W}}^T_k\bm{x}+\bm{\tilde{b}}_k)\right), \nonumber \\
	\bm{y} &= f(\bm{\tilde{W}}^T(\bm{x})\bm{x}+\bm{\tilde{b}}(\bm{x})) \nonumber \\
	\bm{\tilde{W}}(\bm{x})&=\sum_{k=1}^K\pi_k(\bm{x})\bm{\tilde{W}}_k, \:
	\bm{\tilde{b}}(\bm{x})=\sum_{k=1}^K\pi_k(\bm{x})\bm{\tilde{b}}_k \nonumber \\
	\text{s.t.} \;\;\; &0 \leq \pi_k(\bm{x}) \leq 1, 
	\sum_{k=1}^K \pi_k(\bm{x}) = 1,
	\label{eq:dynamic-filter}
\end{align}
where $\pi_k(\bm{x})$ denotes the attention weight for the $k^{th}$ linear function $\bm{\tilde{W}}^T_k\bm{x}+\bm{\tilde{b}}_k$, it is derived from an attention mechanism~\cite{hu2018squeeze} and varies for each input $\bm{x}$. The aggregated weight $\bm{\tilde{W}}(\bm{x})$ and bias $\bm{\tilde{b}}(\bm{x})$ are input-dependent. $\{\bm{\tilde{W}}_k,\bm{\tilde{b}}_k\}$ represent the $k^{th}$ convolution kernel parameters. The combined model $\bm{\tilde{W}}^T(\bm{x})\bm{x}+\bm{\tilde{b}}(\bm{x})$ exhibits non-linear characteristics due to the attention weights, thereby endowing the dynamic perceptron with greater representational capabilities than its static counterpart. Moreover, since these kernels are assembled differently for different inputs based on their input-dependent attention, they can more flexibly recognize and perceive the spatial-agnostic dominant degradations across various input samples.

\begin{table*}[t]
	\centering
	\caption{Quantitative comparison with SOTA SR methods with Setting 1. The best two results are highlighted in {\color{red}{red}} and {\color{blue}{blue}} colors, respectively.} 
	\resizebox{\linewidth}{!}{
		\begin{tabular}{cccccccccccc} 
%			\hline
			\toprule[0.15em]
			\multirow{2}{*}{Method} & \multirow{2}{*}{Scale} & \multicolumn{2}{c}{Set5} & \multicolumn{2}{c}{Set14} & \multicolumn{2}{c}{BSD100} & \multicolumn{2}{c}{Urban100} & \multicolumn{2}{c}{Manga109} \\
			&                        & ~~PSNR~~       & ~~SSIM~~        & ~~PSNR~~        & ~~SSIM~~        & ~~PSNR~~        & ~~SSIM~~         & ~~PSNR~~         & ~~SSIM~~          & ~~PSNR~~         & ~~SSIM~~          \\ 

			\midrule[0.1em]
		%	& \multirow{7}{*}{3}
			Bicubic  & \multirow{7}{*}{2}  & 28.82 &0.8577   &26.02 &0.7634  &25.92 &0.7310	&23.14 &0.7258  &25.60 &0.8498  \\
			
			CARN~\cite{ahn2018fast} & & 30.99 &0.8779	&28.10  &0.7879	   & 26.78 & 0.7286	    &25.27 &0.7630	    & 26.86 &0.8606  \\
			
			Bicubic+ZSSR~\cite{shocher2018zero} & & 31.08 &0.8786	&28.35  &0.7933	   & 27.92 & 0.7632	    &25.25 &0.7618	    & 28.05 &0.8769  \\
			
			Deblurring~\cite{pan2017deblurring}+CARN~\cite{shocher2018zero} & & 24.20 &0.7496	&21.12  &0.6170	   & 22.69 & 0.6471	    &18.89 &0.5895	    & 21.54 &0.7946  \\
			
			CARN~\cite{shocher2018zero}+Deblurring~\cite{pan2017deblurring} & & 31.27 &0.8974	&29.03  &0.8267	   & 28.72 & 0.8033	    &25.62 &0.7981	    & 29.58 &0.9134  \\
			
			IKC~\cite{gu2019blind} & & 37.19 &0.9526	&32.94  &0.9024	   & 31.51 & 0.8790	    &29.85 &0.8928	    & 36.93 &0.9667  \\
			
			DANv1~\cite{luo2020unfolding} & & 37.34 &0.9526 	&33.08  &0.9041	   & 31.76 & 0.8858	    &30.60 &0.9060	    & 37.23 &0.9710  \\
			
			DANv2~\cite{luo2021endtoend} &  & {\color{blue} 37.60} & {0.9544}     & {33.44} & {0.9094}    & {32.00} & {0.8904}   & {31.43} & {0.9174}    & {38.07} & {0.9734}  \\ 
			
                DCLS~\cite{luo2022deep} &  & {\color{red} 37.63} & {\color{red}0.9554}     & {\color{blue}33.46} & {\color{red} 0.9103}    & {\color{blue} 32.04} & {\color{red} 0.8907}   & {\color{blue} 31.69} & {\color{red} 0.9202}    & {\color{red} 38.31} & {\color{red}0.9740}  \\

			GLDFN(Ours) &  & {37.47} & {\color{blue}0.9545}     & {\color{red}33.51} & {\color{blue} 0.9097}    & {\color{red} 32.07} & {\color{red} 0.8907}   & {\color{red} 31.73} & {\color{blue} 0.9200}    & {\color{blue} 38.26} & {\color{blue}0.9737}  \\

			\midrule
			
			Bicubic & \multirow{7}{*}{3}  & 26.21 &0.7766   &24.01 &0.6662  &24.25 &0.6356	&21.39 &0.6203  &22.98 &0.7576  \\
			
			CARN~\cite{ahn2018fast} & & 27.26 &0.7855	&25.06  &0.6676	   & 25.85 & 0.6566	    &22.67 &0.6323	    & 23.85 &0.7620  \\
			
			Bicubic+ZSSR~\cite{shocher2018zero} & & 28.25 &0.7989	&26.15  &0.6942	   & 26.06 & 0.6633	    &23.26 &0.6534	    & 25.19 &0.7914  \\
			
			Deblurring~\cite{pan2017deblurring}+CARN~\cite{shocher2018zero} & & 19.05 &0.5226	&17.61  &0.4558	   & 20.51 & 0.5331	    &16.72 &0.5895	    & 18.38 &0.6118  \\
			
			CARN~\cite{shocher2018zero}+Deblurring~\cite{pan2017deblurring} & & 30.31 &0.8562	&27.57  &0.7531	   & 27.14 & 0.7152	    &24.45 &0.7241	    & 27.67 &0.8592  \\
			
			IKC~\cite{gu2019blind} & & 33.06 &0.9146	&29.38  &0.8233	   & 28.53 & 0.7899	    &24.43 &0.8302	    & 32.43 &0.9316  \\
			
			DANv1~\cite{luo2020unfolding} & & 34.04 &0.9199 	&30.09  &0.8287	   & 28.94 & 0.7919	    &27.65 &0.8352	    & 33.16 &0.9382  \\			
			
			DANv2~\cite{luo2021endtoend} &  & {\color{blue} 34.12} & {\color{blue} 0.9209}     & {30.20} & {0.8309}    & {29.03} & {0.7948}   & {27.83} & {0.8395}    & {33.28} & {0.9400}  \\ 
			
                DCLS~\cite{luo2022deep} &  & {\color{red} 34.21} & {\color{red}0.9218}     & {\color{red}30.29} & {\color{red} 0.8329}    & {\color{red} 29.07} & {\color{blue} 0.7956}   & {\color{red} 28.03} & {\color{red} 0.8444}    & {\color{red} 33.54} & {\color{red}0.9414}  \\
   
			GLDFN(Ours) &  & {33.86} & {0.9203}     & {\color{blue}30.23} & {\color{blue} 0.8317}    & {\color{red} 29.07} & {\color{red} 0.7959}   & {\color{blue} 27.89} & {\color{blue} 0.8423}    & {\color{blue} 33.32} & {\color{blue}0.9405}  \\

			\midrule
			
			Bicubic  & \multirow{7}{*}{4}  & 24.57 &0.7108   &22.79 &0.6032  &23.29 &0.5786	&20.35 &0.5532  &21.50 &0.6933  \\
			
			CARN~\cite{ahn2018fast} & & 26.57 &0.7420	&24.62  &0.6226	   & 24.79 & 0.5963	    &22.17 &0.5865	    & 21.85 &0.6834  \\
			
			Bicubic+ZSSR~\cite{shocher2018zero} & & 26.45 &0.7279	&24.78  &0.6268	   & 24.97 & 0.5989	    &22.11 &0.5805	    & 23.53 &0.7240  \\
			
			Deblurring~\cite{pan2017deblurring}+CARN~\cite{shocher2018zero} & & 18.10 &0.4843	&16.59  &0.3994	   & 18.46 & 0.4481	    &15.47 &0.3872	    & 16.78 &0.5371  \\
			
			CARN~\cite{shocher2018zero}+Deblurring~\cite{pan2017deblurring} & & 28.69 &0.8092	&26.40  &0.6926	   & 26.10 & 0.6528	    &23.46 &0.6597	    & 25.84 &0.8035  \\
			
			IKC~\cite{gu2019blind} & & 31.67 &0.8829	&28.31  &0.7643	   & 27.37 & 0.7192	    &25.33 &0.7504	    & 28.91 &0.8782  \\
			
			DANv1~\cite{luo2020unfolding} & & 31.89 &0.8864 	&28.42  &0.7687	   & 27.51 & 0.7248	    &25.86 &0.7721	    &{30.50} & {0.9037}  \\
			
			DANv2~\cite{luo2021endtoend} &  & {\color{blue} 32.00} & {\color{blue} 0.8885}     & {\color{blue} 28.50} & {0.7715}    & {27.56} & {0.7277}   & {25.94} & {0.7748}    & 30.45 & {\color{blue} 0.9037}  \\ 
			
			AdaTarget~\cite{jo2021adatarget}& & 31.58 &0.8814 	&28.14  &0.7626	   & 27.43 & 0.7216	    &25.72 &0.7683	    & 29.97 &0.8955  \\
   
			DCLS~\cite{luo2022deep} &  & {\color{red} 32.12} & {\color{red}0.8890}     & {\color{red}28.54} & {\color{red} 0.7728}    & {\color{red} 27.60} & {\color{red} 0.7285}   & {\color{red} 26.15} & {\color{red} 0.7809}    & {\color{red} 30.86} & {\color{red}0.9086}  \\
   
			GLDFN(Ours) &  & {31.90} & {0.8869}     & {28.47} & {\color{blue} 0.7719}    & {\color{blue} 27.57} & {\color{blue} 0.7278}   & {\color{blue} 26.00} & {\color{blue} 0.7775}    & {\color{blue} 30.67} & {\color{blue}0.9052}  \\

			\bottomrule[0.15em]
   \label{setting1}
\end{tabular}}
\end{table*}

\begin{table*}[ht]

	\centering
	\caption{Quantitative comparison on various noisy datasets. The best two results are highlighted in {\color{red}{red}} and {\color{blue}{blue}} colors, respectively.}
	\label{setting1_noise}
	\resizebox{\linewidth}{!}{
		\begin{tabular}{cccccccccccc}
			\toprule[0.15em]
			\multirow{2}{*}{Method $\times$4}& \multirow{2}{*}{Noise level} & \multicolumn{2}{c}{Set5~\cite{bevilacqua2012low}}              & \multicolumn{2}{c}{Set14~\cite{zeyde2010single}}     & \multicolumn{2}{c}{BSD100~\cite{martin2001database}}      & \multicolumn{2}{c}{Urban100~\cite{huang2015single}}          & \multicolumn{2}{c}{Manga109~\cite{matsui2017sketch}}                     \\ 
			& \multirow{-2}{*}{}  & PSNR  & SSIM         & PSNR  & SSIM      & PSNR  & SSIM         & PSNR  & SSIM      & PSNR  & SSIM   \\ \midrule[0.1em]
			
			Bicubic+ZSSR~\cite{shocher2018zero} &\multirow{5}{*}{15} & 23.32 &0.4868	&22.49  &0.4256	   & 22.61 & 0.3949	    &20.68 &0.3966	    & 22.04 &0.4952  \\
			IKC~\cite{gu2019blind} & & 26.89 &0.7671	&25.28  &0.6483	   & 24.93 & 0.6019	    &22.94 &0.6362	    & 25.09 &0.7819  \\
			DANv1~\cite{luo2020unfolding} & & 26.95 &0.7711 	&25.27  &0.6490	   & {24.95} & {0.6033}	    &23.00 &0.6407	    &25.29 & 0.7879  \\
			
			DANv2~\cite{luo2021endtoend} &  & {26.97} & {0.7726}     & {25.29} & {0.6497}    & {24.95} & 0.6025   & {23.03} & {0.6429}    & {25.32} & {0.7896}  \\ 
			
			DCLS~\cite{luo2022deep} &  & {\color{blue} 27.14} & {\color{blue} 0.7775}     & {\color{blue} 25.37} & {\color{blue} 0.6516}    & {\color{blue} 24.99} & {\color{blue} 0.6043}   & {\color{blue} 23.13} & {\color{blue} 0.6500}    & {\color{blue} 25.57} & {\color{blue} 0.7969}  \\

               GLDFN(Ours) &  & {\color{red} 27.22} & {\color{red} 0.7798}     & {\color{red} 25.38} & {\color{red} 0.6528}    & {\color{red}  25.02} & {\color{red} 0.6061}   & {\color{red} 23.25} & {\color{red} 0.6541}    & {\color{red} 25.67} & {\color{red} 0.8008}  \\
   
			\midrule
			
			Bicubic+ZSSR~\cite{shocher2018zero} &\multirow{5}{*}{30} & 19.77 &0.2938	&19.36  &0.2534	   & 19.43 & 0.2308	    &18.32 &0.2450	    & 19.25 &0.3046  \\
			IKC~\cite{gu2019blind} & & 25.27 &0.7154	&24.15  &0.6100	   & {24.06} & 0.5674	    &22.11 &0.5969	    & 23.80 &0.7438  \\
			DANv1~\cite{luo2020unfolding} & & 25.32 & {0.7276} 	&24.15  & {\color{blue}0.6138}	   & 24.04 & 0.5678	    &22.08 &0.5977	    &23.82 & 0.7442  \\
			
			DANv2~\cite{luo2021endtoend} &  & { 25.36} &  0.7264     & { 24.16} & 0.6121    & {24.06} &  {0.5690}   & {22.14} & {0.6014}    & {23.87} & {0.7489}  \\ 
			
			DCLS~\cite{luo2022deep} &  & {\color{blue} 25.49} & {\color{blue} 0.7323}     & {\color{blue} 24.23} & {0.6131}    & {\color{red} 24.09} & {\color{blue} 0.5696}   & {\color{red} 22.37} & {\color{blue} 0.6119}    & {\color{red} 24.21} & {\color{red} 0.7582}  \\
			
			GLDFN(Ours) &  & {\color{red} 25.52} & {\color{red} 0.7334}     & {\color{red} 24.24} & {\color{red} 0.6154}    & {\color{red} 24.09} & {\color{red} 0.5715}   & {\color{blue}22.32} & {\color{red} 0.6120}    & {\color{blue}  24.05} & {\color{blue} 0.7578}  \\

			\bottomrule[0.15em]
		\end{tabular}
	}

\end{table*}

\vspace{4pt}
\noindent \textbf{Local Dynamic Filter Layer. }We introduce a local dynamic filter layer for spatial-specific dominant degradations. Compared to standard convolution, the local dynamic filter layer perceives non-uniform degradation types better and enables more flexible handling of localized degradation. We incorporate decoupled dynamic filter~\cite{zhou2021decoupled} into our architecture to build local dynamic filtering layers. 
The local dynamic filter layer features a trifurcated architecture, incorporating identity mapping alongside channel and spatial attention mechanisms, facilitating a richer and more comprehensive capture of feature representations. Applying $1\times1$ convolutions to replace the original fully connected layer can reduce the computational complexity of these two attention mechanisms. The local dynamic filtering is defined as follows:
\vspace{-1mm}
\begin{equation}
	{F'}_{(r,i)} = \sum_{j\in \Omega(i)}{D^{sp}_{i}[\mathbf{p}_i-\mathbf{p}_j] D^{ch}_{r}[\mathbf{p}_i-\mathbf{p}_j] F_{(r,j)}},
	\label{eq:ddconv}
	\vspace{-2mm}
\end{equation}
where ${F'}_{(r,i)}\in\mathbb{R}$ indicates the resulting feature value at the $i^{th}$ pixel and $r^{th}$ channel, ${F}_{(r,j)} \in \mathbb{R}$ is the input feature value at the $j^{th}$ pixel and $r^{th}$ channel.
$\Omega(i)$ represents the $k \times k$ convolution window centered around the $i^{th}$ pixel. The position offset between $i^{th}$ and $j^{th}$ pixels $[\mathbf{p}_i-\mathbf{p}_j] \in \{(-\frac{(k-1)}{2},-\frac{(k-1)}{2}), (-\frac{(k-1)}{2}, -\frac{(k-1)}{2}+1), ..., (\frac{(k-1)}{2},\frac{(k-1)}{2})\}$ where $\mathbf{p}_{i}$ is 2D pixel coordinates. 
$D^{sp} \in \mathbb{R}^{n \times k \times k}$ is the spatial dynamic filter and $D_i^{sp} \in \mathbb{R}^{k \times k}$ represents the filter at $i^{th}$ pixel. $D^{ch} \in \mathbb{R}^{c \times k \times k}$ corresponds to the channel dynamic filter, where$D_r^{ch} \in \mathbb{R}^{k \times k}$ is the filter associated with the $r^{th}$ channel.
As illustrated in Figure~\ref{fig: architecture}, the bottom left corner illustrates the local dynamic filtering module. They leverage the channel and spatial attention mechanisms to generate dynamic filters specific to different channels and spatial locations. We compute the output feature map using the Eq.~\ref{eq:ddconv}. 
Compared to the computational demand of $nc'ck^2$ associated with traditional convolutional layer, the local dynamic filtering module requires only the computational load $nk^2$ plus $ck^2$. $c$ and $c'$ represent the input and output channels, respectively. $n$ represents the number of feature pixels $n = h \times w$, $h$ and $w$ are the width and height of the feature map. Furthermore, this module is trainable and implemented with CUDA acceleration.

The spatial and channel dynamic filters generated by the local dynamic filter layer can be combined into region-specific filters within the image or feature to deal flexibly with the spatial-specific dominant degradations.

\begin{figure*}[t]
	\setlength{\abovecaptionskip}{-0.05in}
	\setlength{\belowcaptionskip}{-0.05in}
	\begin{center}
		\includegraphics[width=.999\linewidth]{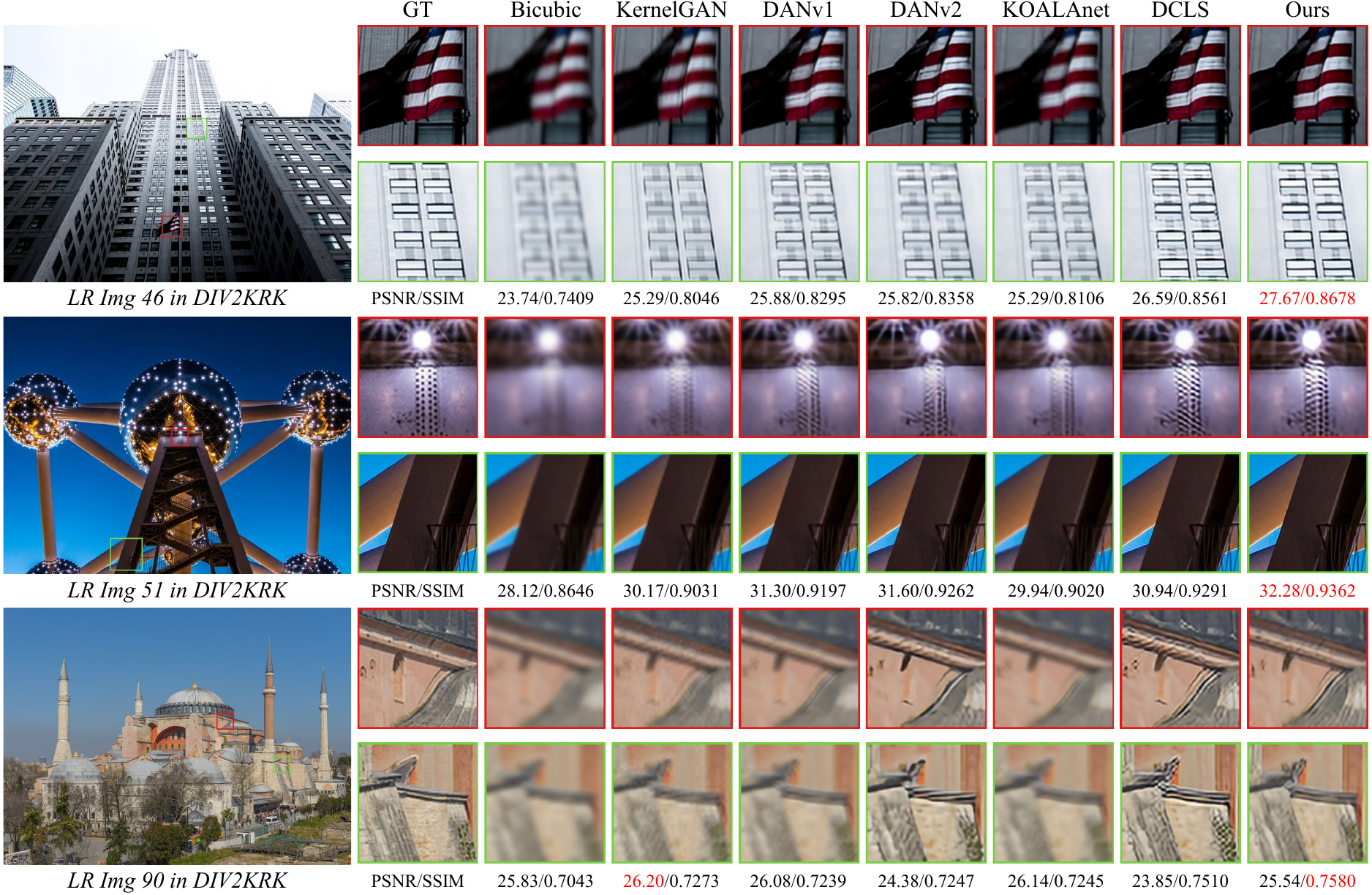}
	\end{center}
	\caption{Visual results of \textit{Img 26},  \textit{Img 37} and \textit{Img 46} in DIV2KRK~\cite{bell2019blind}, for scale factor 2. Please zoom in for the best view.}
	\label{fig:div2krk_x2}
\end{figure*}

\begin{figure*}[t]
	\setlength{\abovecaptionskip}{-0.05in}
	\setlength{\belowcaptionskip}{-0.05in}
	\begin{center}
		\includegraphics[width=.999\linewidth]{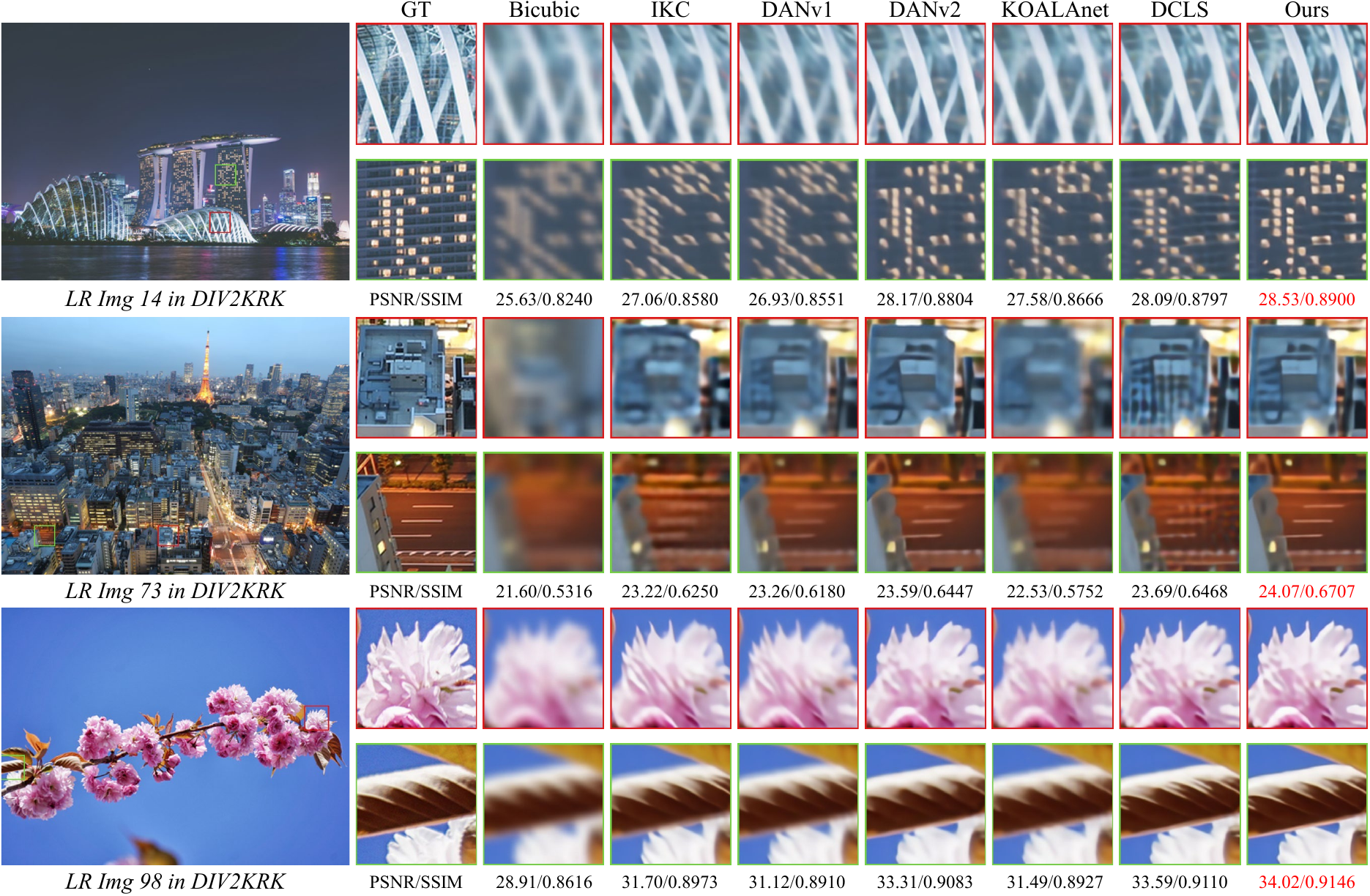}
	\end{center}
	\caption{Visual results of \textit{Img 14},  \textit{Img 73} and \textit{Img 98} in DIV2KRK~\cite{bell2019blind}, for scale factor 4. Please zoom in for the best view.}
	\label{fig:div2krk_x4}
\end{figure*}

\begin{figure*}[t]
	\setlength{\abovecaptionskip}{-0.05in}
	\setlength{\belowcaptionskip}{-0.05in}
	\begin{center}
		\includegraphics[width=.999\linewidth]{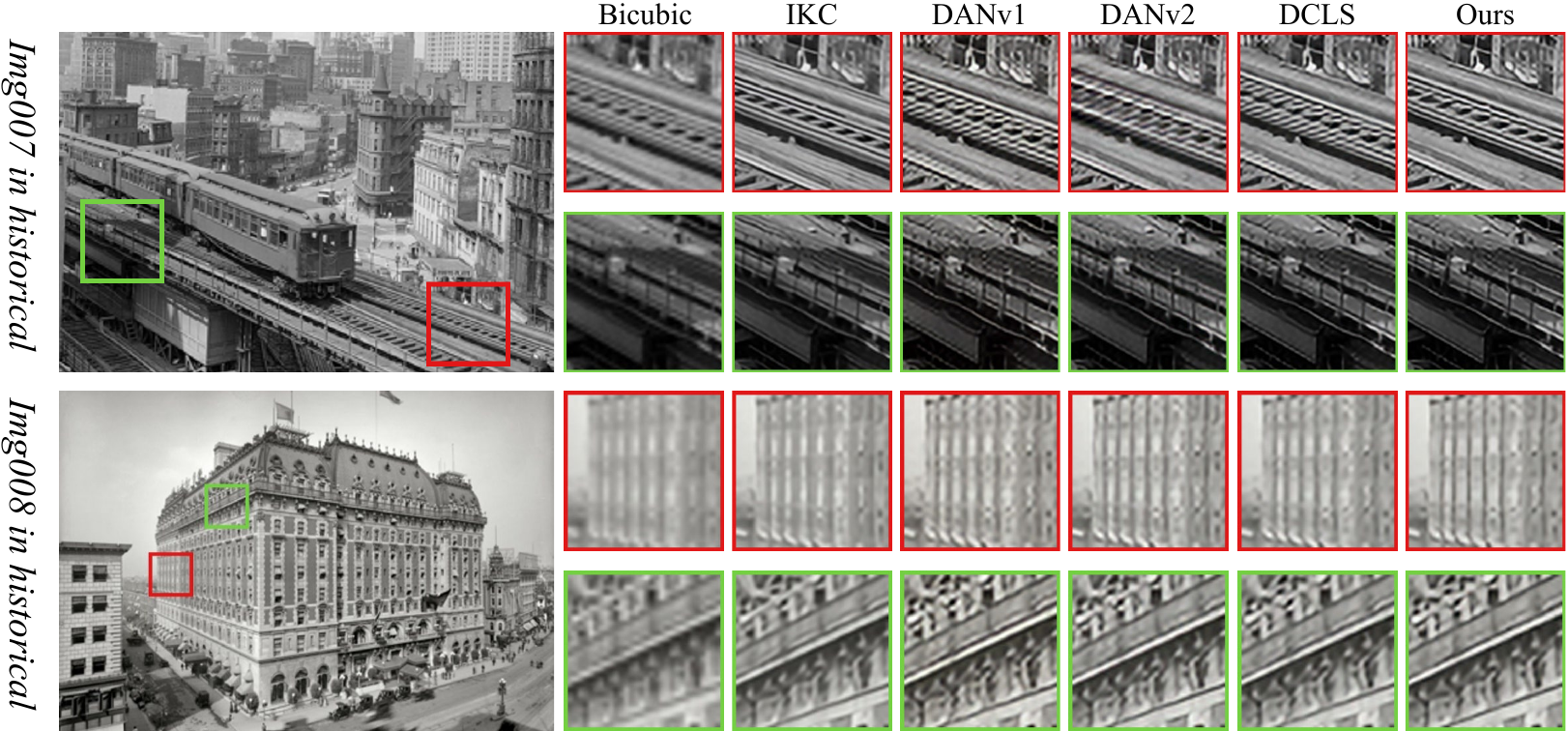}
	\end{center}
	\caption{Visual results of \textit{Img 007} and \textit{Img 008} in Historical~\cite{lai2017deep} dataset, for scale factor 2. Please zoom in for the best view.}
	\label{fig:history_x2}
\end{figure*}

\subsection{Reconstruction Network}
Our experiment found that the interaction of feature information between different branches is conducive to recovering better results, so we introduced a fusion module between the two branches. We first added fusion after each dual dynamic filter module. We found that the feature maps of the two branches tend to be consistent, indicating that the information interaction is too dense, resulting in information redundancy. Therefore, we introduce the fusion module after each dual dynamic filter group (DDFG). As a result, the information of the two branches is different, and the final result is improved. Finally, to save computing resources, we only let the global branch information flow to the local branch and finally form the DDFG shown in Figure~\ref{fig: architecture}.

Upon merging the feature information from both branches through the cascaded DDFGs, we employ the reconstruction module to generate the SR result. This module comprises a convolutional layer and a sub-pixel convolutional layer \cite{shi2016real}. The sub-pixel convolutional layer facilitates scale sampling through pixel translation according to a specified magnification factor. Furthermore, we incorporate a long skip connection to transmit shallow information to the deeper layers,

\begin{equation}
        I^{SR} =\operatorname{Add} \left(\operatorname{U_{\uparrow}}\left(\operatorname{Conv}\left(F_{DDFG}\right)\right),I_{\uparrow}^{{LR}}\right), 
\end{equation}
where $I^{SR}$ denotes the output image of the reconstruction network, and $F_{DDFG}$ represents the output features of the dual dynamic filters refinement network. $ U_\uparrow $ is the operation of sub-pixel convolution, ${I}_{\uparrow}^{LR}$ represents the up-sampled image from $I^{LR}$ using bilinear interpolation.

To optimize our algorithm, we adopt the $L1$ loss to measure the difference between reconstructed images $I^{SR}$ and ground-truth images $ I^{GT} $, which is defined as:
\begin{equation}
	\mathcal{L}_{s r}=\sqrt{\left\|I^{SR}-I^{GT}\right\|^{2}},
\end{equation}
It is worth mentioning that we abstain from employing any additional loss functions, such as adversarial loss \cite{wang2018esrgan,zhang2019image}, perceptual loss \cite{johnson2016perceptual}, and smoothness loss \cite{wang2019learning}.

\section{Experiments}
In this section, we initially present the training datasets and discuss implementation details. Subsequently, we compare the proposed algorithm against several state-of-the-art methods on the benchmark datasets.

\subsection{Training datasets}
We have chosen a total of 3450 HR images from DIV2K~\cite{agustsson2017ntire} and Flickr2K~\cite{timofte2017ntire} datasets to serve as our training dataset, following the approach of~\cite{gu2019blind}. We generate LR images for training and testing based on the principles outlined in Eq.~\ref{eqn: blind}. We assess the performance of our method by evaluating the peak signal-to-noise ratio (PSNR) and structural similarity index (SSIM) on the luminance channel of the super-resolved results in the YCbCr color space. Additionally, we utilize the no-reference evaluation metric, NIQE~\cite{mittal2012making}, on historical datasets to compare the naturalness and quality of the reconstruction results achieved by several SOTA methods.

\vspace{4pt}
\noindent \textbf{Setting 1: Isotropic Gaussian Kernel Setting. }Firstly, we conduct blind SR experiments on isotropic Gaussian kernel setting. Following \cite{gu2019blind,luo2020unfolding}, we uniformly sample the kernel width in [0.2, 2.0], [0.2, 3.0], and [0.2, 4.0] for scale factors $\times2$, $\times3$, $\times4$ SR in the training process. 
To evaluate the effectiveness of our approach, we conducted testing experiments on five widely-used benchmark datasets (Set5~\cite{bevilacqua2012low}, Set14~\cite{zeyde2010single}, BSD100~\cite{martin2001database}, Urban100~\cite{huang2015single}, and Manga109~\cite{matsui2017sketch}).
According to \emph{Gaussian 8} in \cite{luo2020unfolding}, 8 kernels are uniformly chosen from the range [0.80, 1.60], [1.35, 2.40], and [1.80, 3.20] for  $\times2$, $\times3$,  and $\times4$ SR. To obtain the corresponding degraded LR images, we first blur the HR images by \emph{Gaussian 8} and then downsample them with specific scale factors.

\vspace{4pt}
\noindent \textbf{Setting 2: Anisotropic Gaussian Kernel Setting. }To further validate the effectiveness of our method in complex degradation scenarios, we conducted experiments on anisotropic Gaussian kernel settings like~\cite{bell2019blind}.
During the training process, the kernel size is fixed to $11\times11$ and $31\times31$ for scale factors 2 and 4, respectively. We randomly select kernels with a width range in (0.6, 5) and rotation angles in [-$\pi$, $\pi$] to generate anisotropic Gaussian kernels for degradation.
For our testing setting, we choose the DIV2KRK dataset as indicated in \cite{bell2019blind}, which comprises LR images for scale factors $\times2$, $\times4$ SR.

\vspace{4pt}
\noindent \textbf{Setting 3: Spatially Varying Degradation Setting.} We synthesized a test set containing spatial-agnostic and spatial-specific dominant degradations based on the method DMBSR~\cite{laroche2023deep} and named it $COCO\_Valid\_200$. We set two types of noise levels, $[0,5]$ and $[5,10]$, to simulate different degradation conditions. We randomly selected 200 images from the COCO validation set and cropped each image into patches of size $480 \times 480$. Other settings are the same as described in~\cite{laroche2023deep}. We will publicly release this spatially varying test dataset to facilitate future research.

\subsection{Implementation Details}
In our experimental setup, we use five DDFGs, each consisting of ten DDFMs. The input image patch size is $ 64 \times 64 $, and the batch size is 64. We utilize the ADAM optimizer~\cite{kingma2014adam} with $ \beta_{1} = 0.9 $, $\beta_{2} = 0.999$ and $ \epsilon=10^{-8} $. We perform $5 \times 10^5$ iterations on four Nvidia RTX3090 GPUs for approximately three days to train the network for each model and obtain the results presented in our work.
The initial learning rate is set at $4\times10^{-4}$ and halved every $2 \times 10^5$ iterations. Additionally, we apply random rotation and flipping to the data during training for data augmentation.

\begin{figure*}[t]
    \setlength{\abovecaptionskip}{-0.05in}
    \setlength{\belowcaptionskip}{-0.05in}
    \begin{center}
        \includegraphics[width=.99\linewidth]{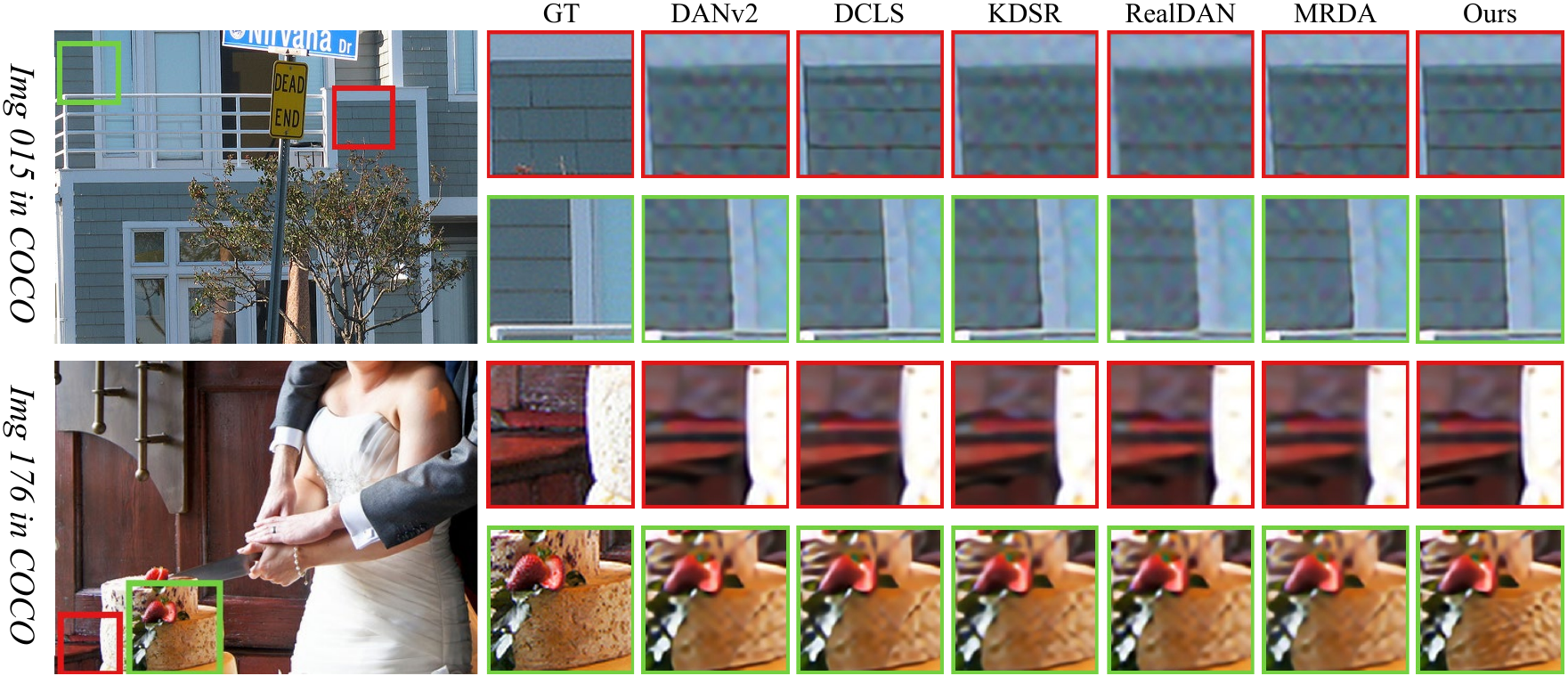}
    \end{center}
        \vspace{4pt}
    \caption{Visual results of \textit{015} and \textit{176} in $COCO\_Valid\_200$ dataset, for scale factor 4. Please zoom in for the best view.}
    \label{fig:coco}
\end{figure*}

\subsection{Ablation Study}
\label{sec-ablation}
This work proposes a dynamic filter-based blind super-resolution network for multiple degradations images. To validate the efficacy of our proposed global and local dynamic filtering modules, we conducted ablation experiments on these crucial components within the proposed approach GLDFN.
To expedite the process, we conducted a mere $2 \times 10^5$ iterations for all ablation experiments (while the other experiments in this paper underwent $5 \times 10^5$ iterations). 
As outlined in Table~\ref{tab: ablation}, we conducted four experiments based on different configurations. In Experiment A, we replaced all global and local dynamic convolution instances in the GLDFN model with standard convolution. Experiment B builds on Experiment A by replacing one convolution branch with a global dynamic filtering module. Similarly, Experiment C builds on Experiment A by replacing a convolutional branch with a local dynamic filtering module. Experiment D is our GLDFN model containing both global and local dynamic filtering layers. Since the parameters of different experiments can vary greatly during the branch replacement process, which is not suitable for directly comparing the performance of different methods, we carefully adjusted the number of modules in the network. This was done to ensure that the number of parameters in Experiments A, B, and C are approximately equal to the number of parameters in Experiment D, thereby facilitating fair comparisons between the experiments.

In Table~\ref{tab: ablation}, for Setting 1 (Urban100), the ablation study results show that local dynamic filters slightly outperform global dynamic filters. This observation contradicts our initial assumption that global dynamic filters would be more effective for isotropic uniform blurring. However, the experimental results indicate a different trend. This discrepancy can be attributed to the unique characteristics of the Urban100 dataset, which includes many urban scenes with rich architectural textures. The proposed local dynamic filter layer, designed to use different filter kernels for pixel blocks in various image regions, offers improved adaptability and flexibility. This design is especially beneficial for processing texture-rich scenes, leading to slightly better performance than the global dynamic filter model.

In the anisotropic Gaussian kernel blur scene (Setting 2), the anisotropic Gaussian blur kernel has different standard deviations along different directions, leading to directionally inhomogeneous blurring. However, this inhomogeneity is spatially invariant across the image. Consequently, global dynamic filters (Experiment B) outperform local dynamic filters (Experiment C) for this type of degradation. Experiment D, which combines global and local dynamic filtering, shows significant improvements, enhancing PSNR by 0.23 dB and SSIM by 0.0045 compared to Experiment A, which only uses convolution. This underscores the effectiveness of our hybrid filtering approach.

In the spatially varying degradation scenario (Setting 3), we synthesized a dataset with various types of degradation, including out-of-focus blur, motion blur, and noise. Given the spatial variability in degradation, local dynamic filtering outperforms global dynamic filtering, consistent with the results in Table~\ref{tab: ablation}. Moreover, global dynamic filtering surpasses Experiment A, indicating its superiority over standard convolution in feature extraction and recovery. Experiment D, which combines global and local dynamic filtering, yields the best results, demonstrating that integrating global and local information enhances performance under complex degradation conditions.

As depicted in Figure~\ref{fig: ablation}, we present the visual outcomes of our ablation experiments. In Experiment A, we employed the convolutional layer as the fundamental component of the dual-branch structure. The third column shows that the convolutional layer can achieve denoising but at the cost of over-smoothing the texture in the restored image. In Experiment B, as shown in the fourth column, we observe a significant improvement in denoising due to introducing the global dynamic filter module. Additionally, owing to the global dynamic filter module's enhanced representational capacity compared to standard convolution, it recovers texture details that closely resemble the high-resolution image. Introducing the local dynamic filtering module in the fifth column enhances image clarity. This improvement is attributed to the local dynamic filtering module's ability to generate region-specific filter kernels based on regional characteristics, enabling more effective handling of degradation types dominated by spatial specificity. Finally, in Experiment D, as shown in the last column, the collaboration between the global and local dynamic filtering modules can achieve more precise and clear reconstruction results. Through our ablation study, we have validated the effectiveness of each module and demonstrated that introducing dynamic filtering modules into the network can better address various image degradation challenges.

\begin{table}[t]
	\small
	\centering
	\setlength\tabcolsep{6pt}
	\caption{Comparisons of Method complexities and inference speed of different methods. Comparisons of different methods NIQE~\cite{mittal2012making} value on the real Historical~\cite{lai2017deep} dataset. The best two results are highlighted in {\color{red}{red}} and {\color{blue}{blue}} colors. The FLOPs are evaluated on an LR size of $180 \times 270$.}
	\begin{tabular}{c ccc}
		\toprule[0.15em]
		Methods                     & Params (M)        & FLOPs (G)       &NIQE$\downarrow$   \\ 
		\midrule[0.15em]
        KernelGAN+ZSSR                  & /			        & /			    & 5.12              \\ 
	%IKC~\cite{gu2019blind}          & 5.29              & 2178.72       & \color{red}4.49   \\
	  DANv1~\cite{luo2020unfolding}	  & 4.33              & 926.72        & 5.13              \\
	  DANv2~\cite{luo2021endtoend}	  & 4.71              & 918.12        & 5.16              \\ 
        DCLS~\cite{luo2022deep}	        & 13.63              & 368.15        & \color{blue}4.87              \\ 
	  GLDFN(Ours)                             & 14.70             & 346.74        & \color{red}4.80   \\ 
		\bottomrule[0.1em]
 \end{tabular}
	\label{tab: params}
\end{table} 

\begin{figure*}[t]
    \setlength{\abovecaptionskip}{-0.05in}
    \setlength{\belowcaptionskip}{-0.05in}
    \begin{center}
        \includegraphics[width=.99\linewidth]{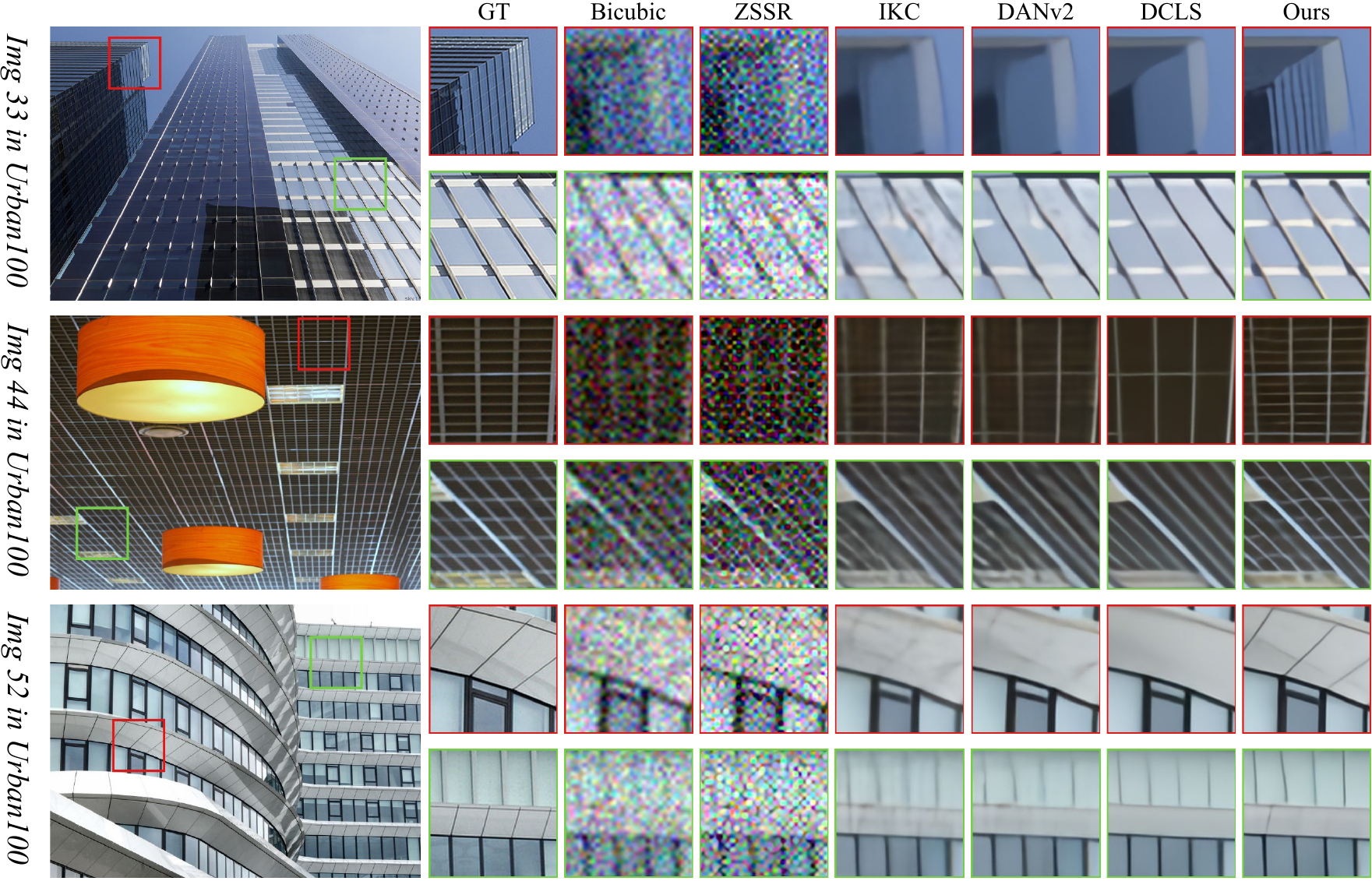}
    \end{center}
        \vspace{4pt}
    \caption{Visual results of \textit{Img 033}, \textit{Img 044} and \textit{Img 052} in Urban100~\cite{huang2015single} dataset, for scale factor 4. Please zoom in for the best view.}
    \label{fig:noise_visual}
\end{figure*}

\subsection{Comparisons with State-of-the-Art Methods}
\noindent \textbf{Setting 1. }In the first setting, we compared several approaches: ZSSR~\cite{shocher2018zero} (with the bicubic kernel), IKC~\cite{gu2019blind}, DANv1~\cite{luo2020unfolding}, DANv2~\cite{luo2021endtoend}, and AdaTargett~\cite{jo2021adatarget}. We also compared CARN~\cite{ahn2018fast} and its variants, implementing blind deblurring method~\cite{pan2017deblurring} before and after CARN.
Following~\cite{gu2019blind}, we synthesized evaluation data using the \emph{Gaussian 8} isotropic blur kernel. We obtained experimental results using official code and pre-trained models for most methods. 
The experimental outcomes, as depicted in Table~\ref{setting1}, reveal that our approach achieves results on par with the state-of-the-art methods in the scenario where the image with only blur degradation.
Additionally, as presented in Table~\ref{setting1_noise}, we introduce varying degrees of noise degradation in the isotropic Gaussian kernel setting. Although our method has a slightly lower PSNR metric than previous methods on individual datasets, it exhibits significant performance improvement on the SSIM metric, which implies our approach can restore better results regarding image brightness, contrast, and structural details. 
As shown in Figure~\ref{fig:noise_visual}, it can be seen that the ZSSR method is unable to handle the noise degradation, and the recovered results contain severe noise; the IKC and DANv2 methods can eliminate the noise effectively, but the results show obvious artifacts; and the DCLS can remove the noise effectively while suppressing the artifacts, but the generated results have smooth transitions and severe loss of structural texture. In contrast, our GLDFN method effectively solves multiple degradation problems, such as blurring and noise, and preserves more texture details. These achievements can be attributed to the robust generalization capabilities of our proposed method, which effectively addresses multiple degradation problems. In contrast, previous blind super-resolution methods, primarily focused on explicit blur kernel estimation, struggled to exhibit superiority across diverse degradation challenges.

\begin{table}[t]
    \setlength{\abovecaptionskip}{0.05in}
    \setlength{\belowcaptionskip}{-0.05in}
    \centering
    \caption{Quantitative comparison on COCO~\cite{lin2014microsoft} in the presence of spatially varying blur and random noise. The best and second best results are shown in {\color{red}{red}} and {\color{blue}{blue}}, respectively.}
    \resizebox{0.9\linewidth}{!}{
        \begin{tabular}{lcccc}
            \toprule[0.15em]
            \multirow{3}{*}{Method}     & \multicolumn{4}{c}{COCO~\cite{lin2014microsoft}}     \\ %\cline{2-5}
            
            & \multicolumn{2}{c}{noise [0,5]}              & \multicolumn{2}{c}{noise [5,10]}    \\ 
            & PSNR  & SSIM         & PSNR  & SSIM  \\ \midrule[0.15em]
                DANv2~\cite{luo2021endtoend}   & 22.83 &0.6275   &22.50  &0.5945  \\
            DCLS~\cite{luo2022deep}    & 22.90 &\color{blue}{0.6305}   &\color{blue}{22.61}  &{\color{blue}{0.6037}}  \\
                RealDAN~\cite{luo2023end} & 22.82 &0.6253   &22.25  &0.6216  \\
            KDSR~\cite{xia2023knowledge}    & 22.90 &0.6302   &22.58  &0.5987  \\
                MRDA~\cite{xia2023meta}   & \color{blue}{22.91} &0.6289   &22.60  &0.5972  \\
                GLDFN(Ours) &{\color{red}{22.93}} &{\color{red}{0.6321}} &{\color{red}{22.68}} &{\color{red}{0.6083}}  \\
                
            \bottomrule[0.1em]
        \end{tabular}
    }
    \label{table:spatial-specific}
\end{table} 

\begin{table}[t]
	\setlength{\abovecaptionskip}{0.05in}
	\setlength{\belowcaptionskip}{-0.05in}
	\centering
 	\caption{Quantitative comparison on DIV2KRK~\cite{bell2019blind}. The best one marks in {\color{red}{red}} and the second best is in {\color{blue}{blue}}.}
	\resizebox{1.0\linewidth}{!}{
		\begin{tabular}{lcccc}
			\toprule[0.15em]
			\multirow{3}{*}{Method}     & \multicolumn{4}{c}{DIV2KRK~\cite{bell2019blind}}     \\ %\cline{2-5}
			
			& \multicolumn{2}{c}{$\times$2}              & \multicolumn{2}{c}{$\times$4}    \\ 
			& PSNR  & SSIM         & PSNR  & SSIM  \\ \midrule[0.15em]
			Bicubic & 28.73 &0.8040   &25.33 &0.6795  \\
			Bicubic+ZSSR~\cite{shocher2018zero} & 29.10 &0.8215   &25.61 &0.6911  \\
			EDSR~\cite{lim2017enhanced} & 29.17 &0.8216   &25.64 &0.6928  \\
			RCAN~\cite{zhang2018image} & 29.20 &0.8223   &25.66 &0.6936  \\
			DBPN~\cite{haris2018deep} & 29.13 & 0.8190   &25.58 &0.6910  \\
			DBPN~\cite{haris2018deep}+Correction~\cite{hussein2020correction} & 30.38 &0.8717   &26.79 &0.7426  \\
			
			KernelGAN~\cite{bell2019blind}+SRMD~\cite{zhang2018learning}  & 29.57 &0.8564   &27.51 &0.7265  \\
			KernelGAN~\cite{bell2019blind}+ZSSR~\cite{shocher2018zero}  & 30.36 & 0.8669   &26.81 &0.7316  \\
			
			IKC~\cite{gu2019blind} & -  & -   &27.70 &0.7668  \\
			DANv1~\cite{luo2020unfolding} & 32.56 & 0.8997      &27.55 &0.7582  \\
			DANv2~\cite{luo2021endtoend} & 32.58 &0.9048     & 28.74 & 0.7893   \\ 
			AdaTarget~\cite{jo2021adatarget} & -  & -   &28.42 &0.7854  \\
			KOALAnet~\cite{kim2021koalanet} & 31.89 & 0.8852     & 27.77 & 0.7637    \\ 
			DCLS~\cite{luo2022deep} & {\color{blue}{32.75}} & {\color{red}{0.9094}}  & {\color{blue}{28.99}} & {\color{blue}{0.7946}}  \\ \midrule
			
			GLDFN(Ours) & {\color{red}{32.88}} & {\color{blue}{0.9084}}  & {\color{red}{29.08}} & {\color{red}{0.7986}}  \\
			\bottomrule[0.1em]
		\end{tabular}
	}

	\label{table:aniso_cmp}
\end{table}
\vspace{4pt}
\noindent \textbf{Setting 2. }In the second configuration, we inject a more challenging anisotropic Gaussian kernel into the training dataset to further verify the method's effectiveness. Anisotropic Gaussian kernels can be seen as a combination of isotropic kernels with motion blur and are more difficult to remove. We use the dataset DIV2KRK~\cite{bell2019blind} containing anisotropic Gaussian kernels for testing.
Similar to \textbf{Setting 1}, we compared several methods: ZSSR~\cite{shocher2018zero} (with the bicubic kernel), IKC~\cite{gu2019blind}, DANv1~\cite{luo2020unfolding}, DANv2~\cite{luo2021endtoend}, AdaTargett~\cite{jo2021adatarget}, KOALAnet~\cite{kim2021koalanet}, and DCLS~\cite{luo2022deep}. Following the work~\cite{luo2022deep}, we compare SOTA SR algorithms designed for bicubic degradation, including RCAN~\cite{zhang2018image} and DBPN~\cite{haris2018deep}. Furthermore, we conducted a comparative analysis involving a two-stage approach that combines the estimation method KernelGAN~\cite{bell2019blind} and the non-blind super-resolution method ZSSR.

We evaluated the performance of various methods on the DIV2KRK~\cite{bell2019blind} dataset. 
Table~\ref{table:aniso_cmp} shows that IKC, DAN, and KOALAnet significantly enhance PSNR and SSIM metrics compared to non-blind SR methods. Moreover, combining these non-blind SR methods with the blur kernel estimation shows an obvious performance improvement. Our method achieves a 0.1db improvement in the PSNR metric compared to recent DCLS methods. Compared to another dynamic filtering-based approach, KOALAnet, our method improves over 1 dB, which proves that our two-branch module designed based on dynamic filtering is effective.

Furthermore, we conducted a visual qualitative comparison experiment to evaluate the effects of different methods. Figure~\ref{fig:div2krk_x2} displays the $\times 2$ SR experimental results on the DIV2KRK dataset, and Figure~\ref{fig:div2krk_x4} presents the results for $\times 4$ magnification reconstruction. In Figure~\ref{fig:div2krk_x2}, both DANv2 and DCLS exhibit more error artifacts, as seen in the white area of the flag in "LR Img 46." These methods produce more noticeable black horizontal bar artifacts, whereas our method yields results closer to HR and avoids such issues. In Figure~\ref{fig:div2krk_x4}, it is evident that recent methods like DANv1, DANv2, and the KOALAnet method struggle to completely remove blur from the image, resulting in a pronounced smearing effect. DCLS performs better than the previous methods in removing the blur but introduces incorrect textures. For instance, in "LR Img 73", certain stripe artifacts, absent in the HR image, appear on roofs and street areas. However, our method produces significantly clearer results and effectively suppresses ringing artifacts.

\vspace{4pt}
\noindent \textbf{Setting 3. }To evaluate the performance of different methods on spatially varying datasets, we compare several state-of-the-art methods on the synthetic dataset $COCO\_Valid\_200$, including DCLS~\cite{luo2022deep}, DANv2~\cite{luo2021endtoend}, MRDA~\cite{xia2023meta}, 
RealDAN~\cite{luo2023end} and KDSR~\cite{xia2023knowledge}, the results are shown in Table~\ref{table:spatial-specific}. From the table, our method improves SSIM by 0.0016 relative to the best method at a noise level of $[0,5]$. It improves by 0.0046 relative to the DCLS method when the noise level increases to $[5,10]$, which indicates that as the noise level increases, the performance gain of our method is greater relative to the other methods, suggesting that our model has better robustness. To better compare the performance of different methods, we show two sets of comparison data, as shown in Figure~\ref{fig:coco}. In the results of \textit{Img 015}, our method is less affected by noise and has clearer lines; in \textit{Img 176}, our method can retain more correct texture details. Our global and local dynamic filtering-based network achieves better noise suppression and texture retention results by comparing qualitative and quantitative results.

\vspace{4pt}
\noindent \textbf{Performance on Real Degradation Images. }To demonstrate the generalization capacity of our approach to realistic degradations, we conduct SR reconstruction experiments on historical images afflicted with compressed artifacts. Neither the ground truth HR images nor blur kernels are available for these images. As depicted in Figure~\ref{fig:history_x2}, we compared the visualization results of our method against several contemporary SOTA methods applied to historical image data. In the first image, "Img 007", we see that DANv1, DANv2, and DCLS will produce unpleasant artifacts when restoring the rails. Although IKC can avoid generating such artifacts, the texture it restores is not sharp enough. Our method can avoid unpleasant artifacts and produce clearer texture details. Our qualitative analysis of the visualizations reveals that our method effectively reduces ringing artifactsringing artifacts compared to DANv1, DANv2, and DCLS, and simultaneously yields sharper results than the IKC method. In Table~\ref{tab: params}, we computed NIQE~\cite{mittal2012making} values for each method. Our proposed approach achieves superior results compared to other methods, which indicates that our method produces images with superior naturalness and quality.
In addition to the above comparison, we have analyzed the parameter count and computational complexity of several recent BSR methods, as depicted in Table~\ref{tab: params}. Due to the loop-based approach used in constructing the network architecture, DAN methods have a relatively smaller parameter count. Both DANv1 and DANv2 employ an iterative strategy to estimate the explicit blur kernel precisely, requiring extensive computations. In contrast, our GLDFN outperforms DANv1 and DANv2 while consuming only $ 38\% $ of the FLOPs required by DANv2.

\section{Conclusions}
\label{sec-con}
We categorize practical degradations into spatial-specific dominant and spatial-agnostic dominant degradations. To address these challenges, we propose a Global and Local Dynamic Filter Network (GLDFN) utilizing local and global dynamic filter layers as fundamental building blocks.
Specifically, we introduce local dynamic filtering layers to address spatial-specific dominant degradations. This layer generates specific filters across different image or feature regions, facilitating the adaptable treatment of diverse degradations in various locations. 
Additionally, we introduce a global dynamic filter layer to handle spatial-agnostic dominant degradations, which generate distinct convolution kernel weights for individual samples. Since the convolution kernel weights are input-dependent, it can effectively discern the spatial-agnostic dominant degradation across samples. 
We pioneer constructing a dual dynamic filtering module using these dynamic filtering layers. Based on this module, we develop a dual dynamic filtering refinement network, effectively resolving spatial-specific dominant and spatial-agnostic dominant degradation problems. Extensive experimental results on synthetic and real-world datasets demonstrate that the proposed model outperforms state-of-the-art blind SR algorithms regarding quantitative results and visual quality.

\section*{Acknowledgment}
This work was supported by the National Natural Science Foundation of China under Grants 62373201 and 61973173; and the Technology Research and Development Program of Tianjin under Grants 18ZXZNGX00340. Computations were supported by the Supercomputing Center of Nankai University.

\ifCLASSOPTIONcaptionsoff
  \newpage
\fi

{
%\small
\bibliographystyle{IEEEtran}
\bibliography{blindsr}
}
\vspace{-15mm}
\begin{IEEEbiography}[{\includegraphics[width=1in,height=1.25in, clip,keepaspectratio]{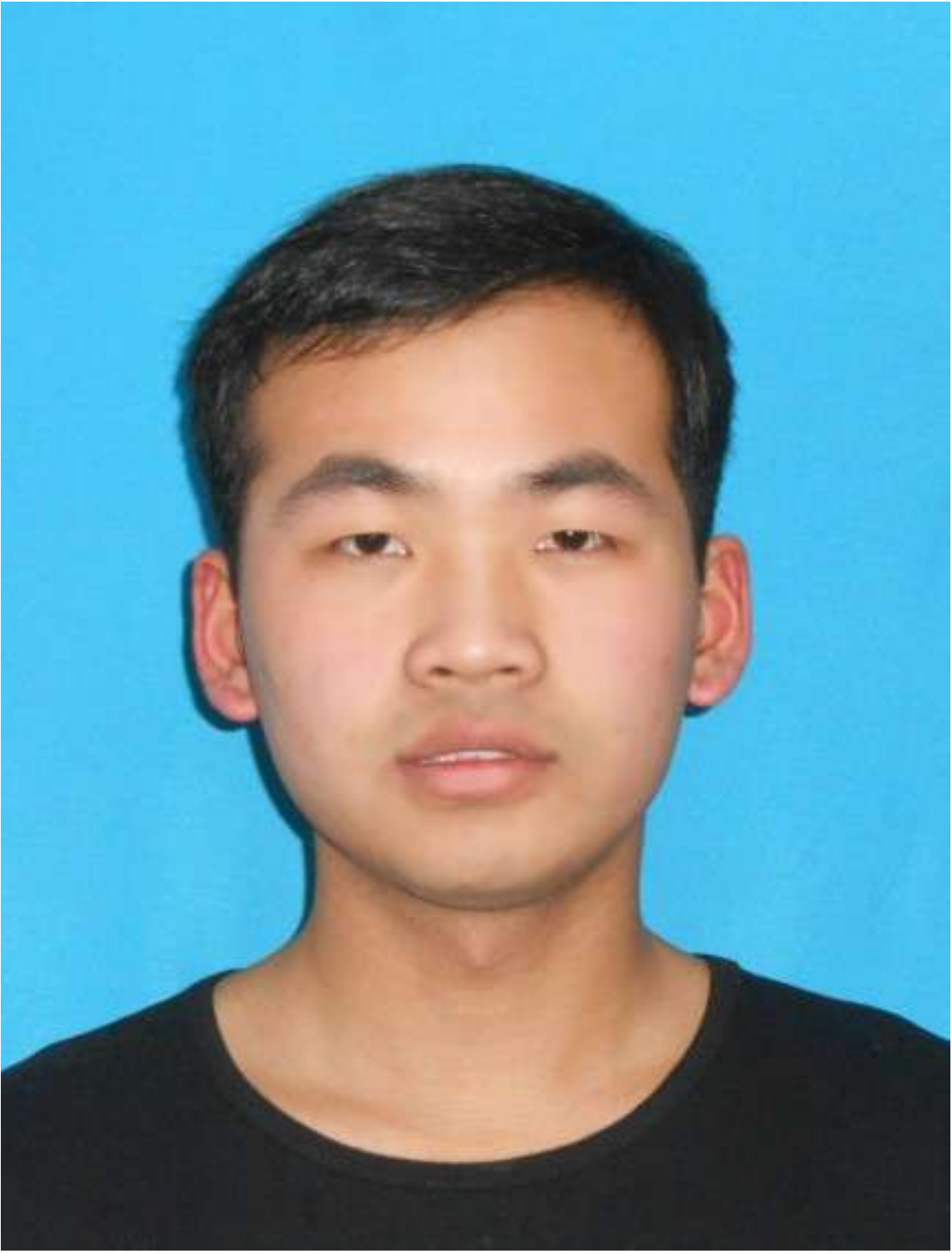}}]{Weilei Wen} received the master's degree from the Xidian University, in 2021. He is currently pursuing a Ph.D. degree at Nankai University, Tianjin, China. His research interests include computer vision and deep learning. He mainly works in the area of generative models and image enhancement. 
\end{IEEEbiography}

\vspace{-15mm}
\begin{IEEEbiography}[{\includegraphics[width=1in,height=1.25in, clip,keepaspectratio]{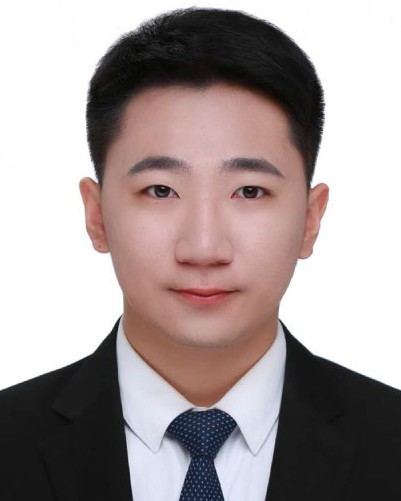}}]{Chunle Guo}(Member, IEEE) received the Ph.D. degree from Tianjin University, China, under the supervision of Prof. Ji-Chang Guo. He was a Visiting Ph.D. Student with the School of Electronic Engineering and Computer Science, Queen Mary University of London (QMUL), U.K. He was a Research Associate with the Department of Computer Science, City University of Hong Kong (CityU of HK). He was a Postdoctoral Researcher with Prof. Ming-Ming Cheng at Nankai University. He is currently an Associate Professor with Nankai University. His research interests include image processing, computer vision, and deep learning.
\end{IEEEbiography}

\vspace{-15mm}
\begin{IEEEbiography}[{\includegraphics[width=1in,height=1.25in, clip,keepaspectratio]
{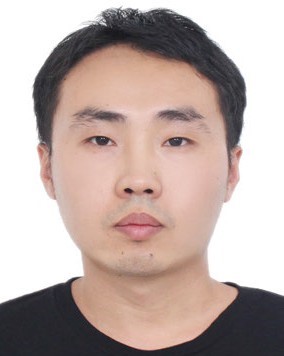}}]{Wenqi Ren}  (Member, IEEE) received the Ph.D. degree from Tianjin University, Tianjin, China, in 2017. From 2015 to 2016, he was supported by the China Scholarship Council and worked with Prof. Ming-Husan Yang as a joint-training Ph.D. student with the Electrical Engineering and Computer Science Department, University of California at Merced. He is currently a Professor with the School of Cyber Science and Technology, Shenzhen Campus, Sun Yat-sen University, Shenzhen, China. His research interests include image processing and related high-level vision problems.
\end{IEEEbiography}

\vspace{-15mm}
\begin{IEEEbiography}[{\includegraphics[width=1in,height=1.25in, clip,keepaspectratio]{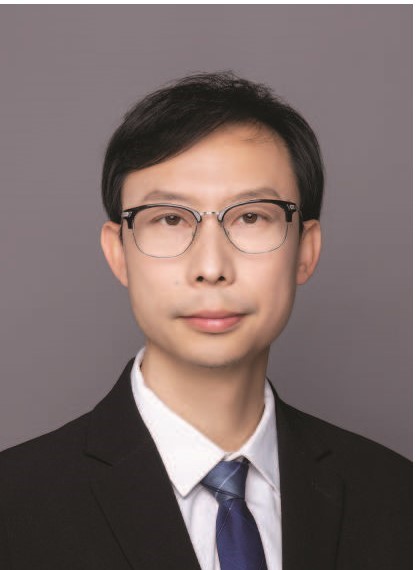}}]
{Hongpeng Wang} (Member, IEEE) received the B.S. degree in physics and the Ph.D. degree in control theory and engineering from Nankai University, Tianjin, China, in 2000 and 2009, respectively. From 2007 to 2008 and from 2016 to 2017, he was a Visiting Research Fellow with the Department of Computer Science and Engineering, Texas A\&M University, College Station, TX, USA. He is currently a Professor with Nankai University. His research interests include artificial intelligence, robotics, virtual reality, and intelligent simulation.
\end{IEEEbiography}

\vspace{-15mm}
\begin{IEEEbiography}[{\includegraphics[width=1in,height=1.25in, clip,keepaspectratio]{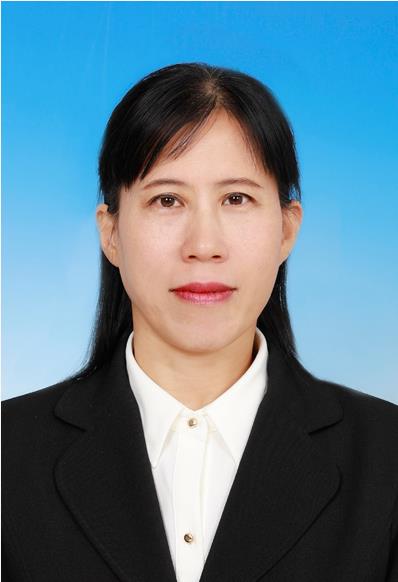}}]{Xiuli Shao} received the B.S. degree in computer software, the M.S. degree in software theory, and the Ph.D. degree in control theory and control engineering from Nankai University, Tianjin, China, in 1986, 1991 and 2002, respectively. She is currently a Professor at the College of Computer Science, Nankai University. Her research interests include artificial intelligence, data analysis, and software engineering. 
	
\end{IEEEbiography}

\end{document}